\documentclass[10pt,twocolumn,letterpaper]{article}

\usepackage[pagenumbers]{cvpr} 

\usepackage[dvipsnames]{xcolor}
\usepackage[utf8]{inputenc} 
\usepackage[T1]{fontenc}    
\usepackage{url}            
\usepackage{booktabs}       
\usepackage{amsfonts}       
\usepackage{amsmath}
\usepackage{nicefrac}       
\usepackage{microtype}      
\usepackage{graphicx}
\usepackage{graphbox}
\usepackage{subcaption}
\usepackage{algorithm}
\usepackage{algorithmic}
\usepackage{bm}
\usepackage[accsupp]{axessibility} 
\usepackage{ulem}
\usepackage{colortbl,array}
\usepackage{multirow}
\usepackage{makecell}

\newcommand{\TODO}[1]{}
\renewcommand{\TODO}[1]{{\color{cyan} [TODO: {#1}]}}
\newcommand{\WIP}[1]{}
\renewcommand{\WIP}[1]{{\color{magenta} [WIP: {#1}]}}
\definecolor{midblue}{rgb}{0,0.11372549,0.258823529}

\newcommand{\argmin}{\mathop{\rm arg~min}\limits}

\definecolor{cvprblue}{rgb}{0.21,0.49,0.74}
\usepackage[pagebackref,breaklinks,colorlinks,citecolor=cvprblue]{hyperref}

\def\paperID{*****} 
\def\confName{CVPR}
\def\confYear{2026}
\newcommand{\proposed}{Parallel-ICL}
\newtheorem{theorem}{Theorem}[section]
\definecolor{LightBlue}{rgb}{0.88,0.92,0.95} 
\definecolor{Blue}{rgb}{0.84,0.84,0.95} 
\definecolor{DeepBlue}{rgb}{0.66,0.66,0.95} 
\definecolor{LightYellow}{rgb}{0.98,0.96,0.89} 
\definecolor{LightRed}{rgb}{0.99,0.91,0.91}    

\title{Parallel In-context Learning for Large Vision Language Models}

\author{%
  Shin'ya Yamaguchi\thanks{Corresponding author. \texttt{shinya.yamaguchi@ntt.com}}\\
  NTT \\
  \and  
  Daiki Chijiwa \\
  NTT \\
  \and
  Tamao Sakao \\
  NTT\\
  \and
  Taku Hasegawa\\
  NTT\\
}

\begin{document}

\maketitle

\begin{abstract}
Large vision-language models (LVLMs) employ multi-modal in-context learning (MM-ICL) to adapt to new tasks by leveraging demonstration examples. While increasing the number of demonstrations boosts performance, they incur significant inference latency due to the quadratic computational cost of Transformer attention with respect to the context length. To address this trade-off, we propose Parallel In-Context Learning (Parallel-ICL), a plug-and-play inference algorithm. Parallel-ICL partitions the long demonstration context into multiple shorter, manageable chunks. It processes these chunks in parallel and integrates their predictions at the logit level, using a weighted Product-of-Experts (PoE) ensemble to approximate the full-context output. Guided by ensemble learning theory, we introduce principled strategies for Parallel-ICL: (i) clustering-based context chunking to maximize inter-chunk diversity and (ii) similarity-based context compilation to weight predictions by query relevance. Extensive experiments on VQA, image captioning, and classification benchmarks demonstrate that Parallel-ICL achieves performance comparable to full-context MM-ICL, while significantly improving inference speed. Our work offers an effective solution to the accuracy-efficiency trade-off in MM-ICL, enabling dynamic task adaptation with substantially reduced inference overhead.
\end{abstract}

\section{Introduction}\label{sec:intro}
Large vision-language models (LVLMs) are text-generative models that integrate pre-trained large language models (LLMs) with vision encoders, processing both text and image inputs~\cite{Tsimpoukelli_NeurIPS21_multimodal_few, Alayrac_NeurIPS22_flamingo, Dai_NeurIPS23_InstructBLIP, liu_NeurIPS23_llava, Achiam_2023_GPT4, Bai_2025_Qwen-vl, Google_2025_gemma3}. 
Through instruction tuning on extensive text-image datasets, LVLMs have demonstrated remarkable success in solving general and complex multi-modal reasoning tasks specified by users~\cite{liu_NeurIPS23_llava, Liu_CVPR24_improved_visual_instruction_tuning}. 
To fully leverage the generalizability of these models on unseen tasks, multi-modal in-context learning (MM-ICL,~\cite{Zhao_ICLR24_mmicl}) has emerged as a crucial paradigm~\cite{Zhang_CVPR25_lvlm_ood}. 
Similar to ICL in LLMs~\cite{brown_NIPS20_gpt3}, MM-ICL conditions the LVLM's decoding on multiple demonstration examples, each consisting of an image-text pair input and its corresponding output, interleaved with the input query.
This contextual guidance enables the LVLM to comprehend the task at hand, allowing it to dynamically adapt to novel tasks and achieve impressive performance during inference without requiring any parameter updates~\cite{Alayrac_NeurIPS22_flamingo, Sun_CVPR24_emu2, Gadetsky_ICLR25_lvlm_unsupervised_icl}.
While the performance gains from MM-ICL typically scale with the number of demonstrations~\cite{Baldassini_CVPRW24_what_makes_mmicl, Qin_NeurIPS24_mmicl_factors, Zhang_CVPR25_lvlm_ood, Zong_ICLR2025_vl-icl_bench}, this presents a significant challenge: most LVLMs are built upon Transformer architectures, which incur a computational cost that scales quadratically with respect to the length of the input context in the worst case~\cite{Vaswani_NeurIPS17_attention_transformer}.
As a result, increasing the number of demonstrations dramatically drops inference speed in exchange for accuracy, as shown in Table~\ref{tb:pre_ex}.
This issue is particularly severe for MM-ICL, as LVLMs often represent images using many visual tokens to capture fine-grained details~\cite{Li_2024_llavanext, Li_arXiv24_llava-ov}, exacerbating the inference overhead associated with longer contexts.

\begin{figure}[t]
    \centering
    \includegraphics[width=1.0\linewidth]{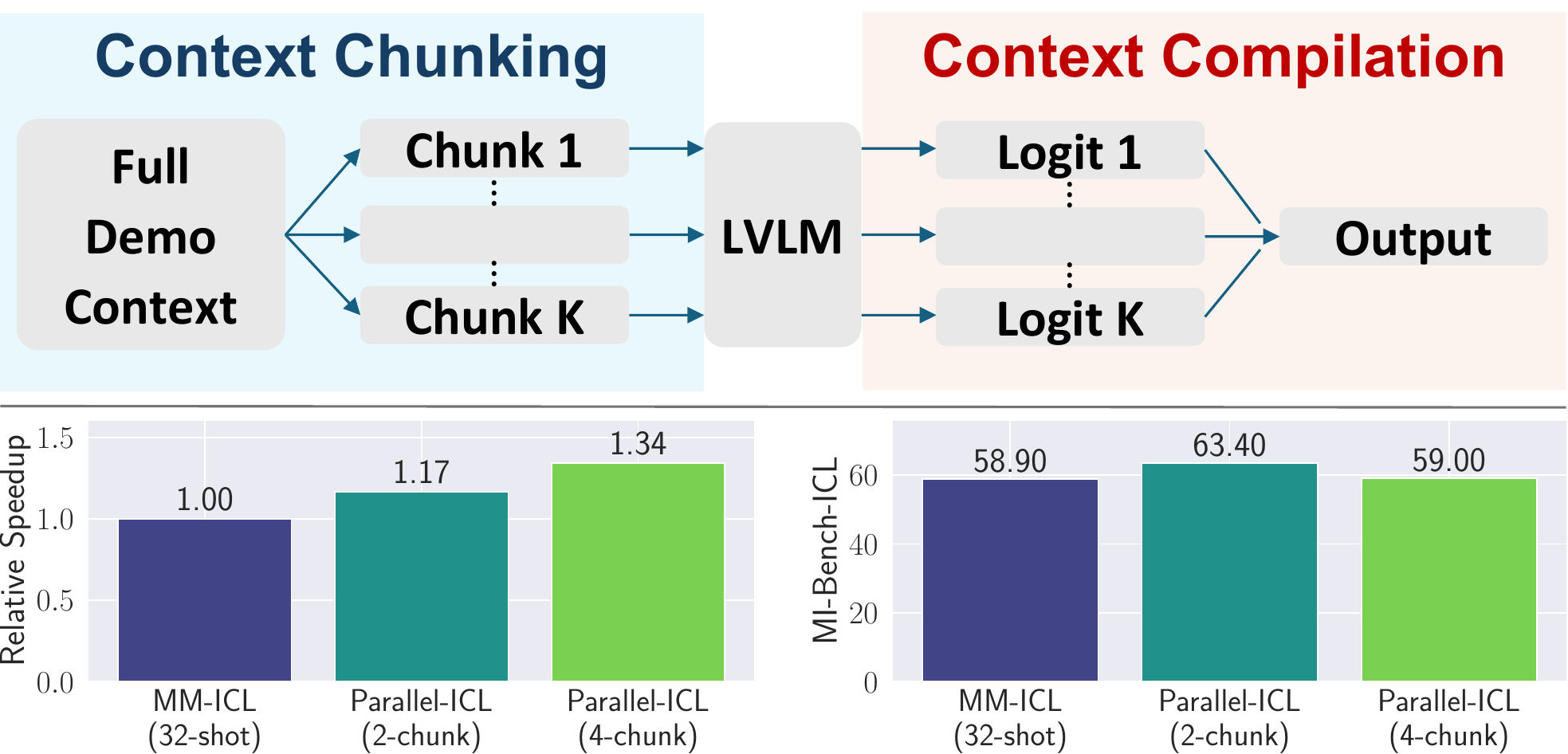}
    \caption{
    \textbf{Parallel-ICL.} 
    Instead of using a full demonstration context, we propose partitioning it into smaller chunk contexts (context chunking) and then integrating the logits from the chunked contexts to output (context compilation) for efficient inference in multi-modal in-context learning (MM-ICL) by large vision-language models (LVLMs).
    Parallel-ICL enhances inference speed while maintaining competitive performance to the original MM-ICL.
    }
    \label{fig:top}
    \vspace{-5mm}
\end{figure}

To reduce the inference cost, prior work has focused on approximating MM-ICL without demonstrations at inference time.
One prominent approach involves extracting a \textit{task vector} from the intermediate activations of the model by processing a large set of demonstrations in advance~\cite{Hendel_EMNLP23_icl_task_vector, Peng_NeurIPS24_live, Zhuang_ICLR25_vectoricl, Jiang_CVPR25_mimic_mmicl}.
However, this method often requires hundreds of demonstrations to achieve performance comparable to that of few-shot MM-ICL, and incurs additional optimization to construct the vectors.
These limitations deviate from the primary goal of MM-ICL, i.e., dynamic and effortless adaptation to new tasks, and impose additional burdens on the user beyond inference.
This leads us to our primary research question: \textit{Can we efficiently approximate long-context MM-ICL at inference time without any additional burdens?}

The major cause of inefficiency in MM-ICL is the long sequential demonstration context. 
However, individual demonstrations, i.e., a tuple of (image, question, answer), are often independent and do not strictly need to be processed as a single series.
Based on this observation, we propose Parallel In-context Learning (\proposed{}), a novel, efficient, and plug-and-play inference algorithm for MM-ICL. 
The core idea is twofold: (i) \textit{context chunking}, where the long demonstration context is partitioned into multiple shorter, manageable chunks, and (ii) \textit{context compilation}, where predictions from each chunk are integrated to approximate the full-context prediction.
We formulate this idea as a weighted Product-of-Experts (PoE) ensemble~\cite{Hinton2002_poe}, approximating the full-context distribution $p(y|\mathcal{C}, x, t)$ as $\prod_{k=1}^{K} p(y|C_k, x, t)^w_i$, where $\mathcal{C}=\{C_1, \dots, C_K\}$, $\mathcal{C}_k$ is the $k$-th chunk, $x$ and $t$ are the input image and query text, and $w_k$ is the weight for the $k$-th chunk.
By the parallel batch-wise computation, \proposed{} effectively ensembles task-specific knowledge from demonstrations with lower overhead than computing the full context at once.

Inspired by the theoretical principles of ensemble methods in machine learning~\cite{Brown_2009_information_theory_ensemble, Zhou_2010_multi-information_ensemble, Morishita_ICML22_rethinking_fano_ensemble}, particularly those related to Fano's inequality~\cite{Fano_1961_transmission_of_information}, we propose principled chunking and compilation strategies designed to maximize the ensemble's effectiveness. 
Our strategies are based on two key factors: inter-chunk diversity and query relevance.
For chunking, we leverage a clustering algorithm to partition the demonstrations. This strategy aims to maximize the diversity among chunks, which is theoretically required for a small prediction error in the ensemble.
For compilation, we assign higher weights to the predictive distributions of chunks that exhibit greater similarity to the input query.
This is expected to enhance the final ensemble prediction.\looseness-1

We conduct extensive experiments on various benchmarks, including image captioning, classification, and visual question answering (VQA), using several state-of-the-art LVLMs, such as LLaVA-OV~\cite{Li_arXiv24_llava-ov}, Qwen2.5-VL~\cite{Bai_2025_Qwen-vl}, and InternVL3.5~\cite{Wang_arXiv25_internvl3.5}. 
Our results demonstrate that the ICL capability of LVLMs emerges even when using an ensemble of next-token predictions with short context chunks, yielding strong accuracy at low latency.
Intriguingly, in several cases, we observe that \proposed{} can outperform the full-context MM-ICL in accuracy.
This suggests that ensembling chunked contexts may mitigate the so-called ``lost in the middle''~\cite{Liu_TACL24_lost_in_the_middle}, which is information loss in MM-ICL when dealing with very long contexts.
Furthermore, we observe that the predictive distribution produced by \proposed{} yields high inter-chunk diversity and query relevance, validating the efficacy of our proposed chunking and compilation strategies.
We believe that our work not only achieves a better accuracy-efficiency trade-off but also introduces a new paradigm that integrates information from multiple and diverse contexts.
This could enable complex multi-modal reasoning even beyond the training time context lengths.

\begin{figure*}[t]
    \centering
    \includegraphics[width=\linewidth]{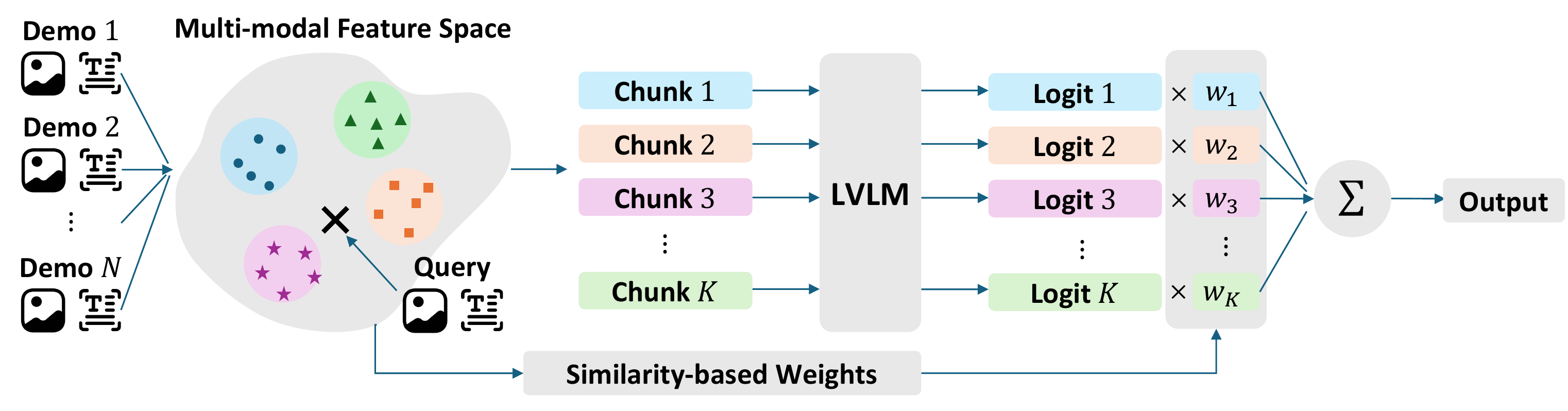}
    \caption{
        \textbf{Pipeline of \proposed}. We cluster demonstrations in a multi-modal feature space and utilize each cluster as chunks (context chunking). 
        Then, we process chunk-wise contexts with LVLMs and weight their outputs (logits) based on query-chunk similarity, composing an ensemble for the final prediction as PoE (context compilation).
        This can be computed by the weighted sum of outputs at the logit level.
    }
    \label{fig:method}
    \vspace{-5mm}
\end{figure*}

\section{Related Work}\label{sec:relatedwork}
\noindent\textbf{Large Vision Language Models (LVLMs).}
LVLMs are text-generative models integrating a pre-trained large language model (LLM) and a visual encoder via a projector to translate visual features into LLM's token embeddings~\cite{Tsimpoukelli_NeurIPS21_multimodal_few, Alayrac_NeurIPS22_flamingo, Dai_NeurIPS23_InstructBLIP, liu_NeurIPS23_llava, Achiam_2023_GPT4, Bai_2025_Qwen-vl, Google_2025_gemma3}.
This integration is often achieved via visual instruction tuning~\cite{liu_NeurIPS23_llava}, which fine-tunes the models on diverse multi-modal tasks, including image captioning, VQA, object detection, and OCR~\cite{Li_2024_llavanext, Lu_arXiv24_deepseek-vl, Liu_CVPR24_improved_visual_instruction_tuning, Zhang_ECCV24_gpt4roi, Bai_2025_Qwen-vl}.
Recent LVLMs have also evolved to process multiple (interleaved) images and video by scaling the model parameters, datasets, and the number of visual tokens for visual instruction tuning~\cite{Li_2024_llavanext, Chen_CVPR24_internvl}.
Similar to LLMs, LVLMs possess zero-shot capability for unseen tasks given by users' input prompts~\cite{liu_NeurIPS23_llava, Achiam_2023_GPT4} and can solve the tasks through the contextual guidance, such as chain-of-thought prompting~\cite{Zhang_TMLR_mm-cot, Mondal_AAAI24_kam-cot, Mitra_CVPR24_CCoT, Gao_CVPR25_interleaved-modal_cot, Xu_ICCV25_llava-cot} and in-context learning~\cite{Alayrac_NeurIPS22_flamingo, Laurenccon_NeurIPS23_obelics_idefics, Zhao_ICLR24_mmicl, Shukor_ICLR24_beyond_task_perf_mmicl}.
In this paper, we assume that LVLMs are trained with instruction tuning that enables in-context learning for unseen multi-modal tasks.

\noindent\textbf{Multi-modal In-context Learning (MM-ICL).}
In-context learning (ICL,~\cite{brown_NIPS20_gpt3}) is an adaptation paradigm for LLMs on unseen tasks, defined by user instructions and multiple demonstrations (input-output pairs)~\cite{Chowdhery_JMLR23_palm, Wu_NeurIPS24_scaling_inference_computation_llm}.
LVLMs inherit the ICL capability and show remarkable performance in multi-modal reasoning tasks~\cite{Laurenccon_NeurIPS23_obelics_idefics, Zhao_ICLR24_mmicl}, especially out-of-distribution tasks~\cite{Zhang_CVPR25_lvlm_ood}.
We refer to this capability for multi-modal tasks as multi-modal ICL (MM-ICL).
For MM-ICL, several studies~\cite{Baldassini_CVPRW24_what_makes_mmicl, Qin_NeurIPS24_mmicl_factors, Chen_WACV25_truly_mmicl} have revealed important demonstration properties, such as their size, selection strategy, and order.
Since longer demonstration contexts degrade inference speed, these studies have also proposed empirical methods for constructing effective input prompts based on demonstration search algorithms with query similarity~\cite{Baldassini_CVPRW24_what_makes_mmicl, Qin_NeurIPS24_mmicl_factors}.
These techniques are also applicable in our method, as the problem setting is compatible with the original MM-ICL.

\noindent\textbf{Efficient Inference for MM-ICL.}
To reduce the inference overhead of MM-ICL, \citet{Peng_NeurIPS24_live} and \citet{Jiang_CVPR25_mimic_mmicl} have leveraged \textit{task vectors}~\cite{Hendel_EMNLP23_icl_task_vector} extracted from intermediate activation outputs of Transformers to internally mimic MM-ICL without using demonstrations at inference time.
In principle, they first detect attention heads where the outputs should be replaced by task vectors via optimizations or heuristics, and then compute task vectors from many demonstrations.
Finally, LVLMs output tokens by replacing the activation vectors with the task vectors.
While these methods can omit demonstrations from input contexts and speed up inference, they require a large number of demonstrations and per-task optimizations beyond inference, compromising the flexibility and dynamic nature of the original MM-ICL.
In contrast, our method is designed in a plug-and-play manner and can be dynamically performed without requiring any additional datasets or optimizations.

\noindent\textbf{Efficient Inference for General LVLMs.}    
Efficient LVLM inference is a primary concern, as an image often requires thousands of tokens to be represented in LVLMs~\cite{Bai_2025_Qwen-vl}.
In this context, the current mainstream can be roughly categorized into two directions: (i) pruning or merging unimportant tokens or KV cache entries to reduce redundancy~\cite{Chen_ECCV24_fastv_lvlm_token_pruning, Liu_ECCV24_elastic_cache, Alvar_CVPR25_divprune, Chen_NeurIPS24_llavolta, Yang_CVPR25_pvc, Zhang_ICML25_sparsevlm, Xing_CVPR25_pyramid_drop}, and (ii) redesigning model architectures for more efficient inference~\cite{Gagrani_CVPRW24_speculative_decoding_for_lvlm, Liang_CVPR25_efficientllava, Huang_ICLR25_dynamic_llava, Zhang_ICLR25_llava-mini, Ji_EMNLP25_specvlm}.
Our research can also be viewed as one approach for efficient LVLM inference, but it differs in that our method is specifically tailored for MM-ICL. 
Furthermore, our method is applicable in a plug-and-play manner and is largely orthogonal to these general efficiency methods, allowing it to be easily combined with them.\looseness-1

\section{Preliminaries}\label{sec:preliminaries}
We briefly introduce formulations of decoding and multi-modal in-context learning (MM-ICL) with large vision language models (LVLMs).
We also demonstrate our motivation regarding the accuracy-efficiency trade-off in MM-ICL through preliminary experiments.

\subsection{Decoding in Large Vision Language Models}
Consider an auto-regressive LVLM parameterized by $\theta$, which is trained on large-scale datasets to accept both images and texts as input for its backbone LLM.
Given an input image $x$ and query text $t$, the probability for yielding the output token sequence $\bm{y}=(y_1,\dots,y_L)\in\mathcal{V}^L$ is defined as
\begin{equation}
     p_\theta(\bm{y}|x,t) = \prod^{L}_{i=1}p_\theta(y_i|\bm{y}_{<i},x,t),\label{eq:lvlm_cond_dist}
\end{equation}
where $L$ is token length, $\mathcal{V}$ is textual token vocabulary, and $\bm{y}_{<i}$ are preceding output tokens.
Similar to LLMs, $p_\theta(y_i|\bm{y}_{<i},x,t)$ over $\mathcal{V}$ is the softmax of the model's output logit ${l}_\theta(y_i|\bm{y}_{<i},x,t)$:
\begin{gather}
     p_\theta(y_i=w|\bm{y}_{<i},x,t) = \operatorname{softmax}(l_\theta(w|\bm{y}_{<i},x,t))
\end{gather}
Each token $y_i$ is generated from $p_\theta(y_i|\bm{y}_{<i},x,t)$ via a decoding strategy such as greedy decoding, i.e., $y_i = \arg \max_{w\in\mathcal{V}} p_\theta(y_i = w|\bm{y}_{<i},x,t)$.
We basically adopt greedy decoding throughout this paper.

\subsection{Multi-modal In-context Learning (MM-ICL)}
Given a set of $N$ demonstrations $\mathcal{E}=\{e^j|e^j=(x^j,t^j,y^j)\}^N_{j=1}$, an LVLM performs MM-ICL for an input image $x$ and query text $t$ by a next-token prediction:
\begin{equation}
     p(\bm{y}|\mathcal{E},x,t) = \prod^{L}_{i=1}p_\theta(y_i|\bm{y}_{<i}, \mathcal{E},x,t),\label{eq:mmicl}
\end{equation}
where $x^j,t^j,y^j$ are the $j$-th image, text question, and ground-truth output for the demonstration.
By adding $\mathcal{E}$ into the input context, LVLMs can comprehend the input-output relations in tasks through the demonstrations, and effectively condition the output prediction via the attention mechanism in Transformer~\cite{brown_NIPS20_gpt3, Zhao_ICLR24_mmicl}.
To produce accurate predictions for unseen tasks, it is crucial to increase the number of demonstrations $N$ as much as possible~\cite{Zhang_CVPR25_lvlm_ood}.
However, since LVLMs require thousands of tokens to represent images, increasing $N$ significantly slows down the inference of MM-ICL as shown in the next section.

\section{Accuracy-Efficiency Trade-off in MM-ICL}
We demonstrate our research motivation by analyzing the accuracy-efficiency trade-off in MM-ICL.
We performed MM-ICL with LLaVA-OV-7B~\cite{Li_arXiv24_llava-ov} using FlashAttention-2~\cite{Dao_ICLR24_flashattention2} on MI-Bench-ICL~\cite{Liu_EMNLP24_mibench} (please see Section~\ref{sec:ex_setting} for more details).
Table~\ref{tb:pre_ex} shows the task accuracy and latency when varying the demonstration number $N$ in 0, 8, 16, and 32.
We observe a clear trade-off: while MM-ICL's performance improves as $N$ increases, the latency significantly grows, even with optimizations like FlashAttention-2.
For a deeper analysis, we also provide a breakdown of prefilling (i.e., pre-computing the KV cache of the input context in advance) and decoding latency.
The results show that prefilling in MM-ICL is dominant for the total inference, indicating that longer input contexts, due to larger demonstration sizes, inevitably degrade inference speed.
This severe trade-off motivates us to develop a new MM-ICL paradigm that improves inference speed by circumventing the need to process the full long input context.

\begin{table}[t]
    \centering
    \caption{\textbf{Analysis of LVLM Inference}. We perform the inference on demo-based task learning of MI-Bench-ICL (demo-based learning)~\cite{Liu_EMNLP24_mibench} with LLaVA-OV-7B~\cite{Li_arXiv24_llava-ov}. Accuracy increases proportionally with the number of demonstrations, while inference time increases significantly.
    }\label{tb:pre_ex}
    \resizebox{\linewidth}{!}{
    \begin{tabular}{lccccc}\toprule
         \multirow{2}{*}{\textbf{Method}} & \multirow{2}{*}{\textbf{\makecell{Token\\Length}}} & \multirow{2}{*}{\textbf{Accuracy}} & \multicolumn{3}{c}{\textbf{Latency (sec)}}\\\cmidrule{4-6}
          &  & & Total & Prefill & Decoding\\\midrule
         Zero-shot        &  2,557 & 0.00  & 0.099 & 0.087 & 0.012 \\
         MM-ICL (8-shot)  & 23,318 & 56.90 & 1.004 & 0.977 & 0.027 \\
         MM-ICL (16-shot) & 44,027 & 58.20 & 2.376 & 2.343 & 0.033 \\
         MM-ICL (32-shot) & 84,959 & 58.90 & 3.479 & 3.444 & 0.035 \\
         \bottomrule
    \end{tabular}
    }
    \label{tab:placeholder}
    \vspace{-5mm}
\end{table}

\section{Method}\label{sec:method}
We propose \proposed{}, an efficient MM-ICL inference method that constructs the next-token distribution for output sequences from chunked demonstration contexts instead of full-length ones (Figure~\ref{fig:method}).
\proposed{} involves two stages: \textit{context chunking} to partition the full contexts, and \textit{context compilation} to aggregate chunk-wise predictive distributions via a logit-level ensemble.
In this section, we formalize the objective of \proposed{}. 
Next, we observe desirable properties for \proposed{} from theoretical perspectives, and introduce our context chunking and compilation algorithms inspired by the observations.

\subsection{Objective}
The objective of \proposed{} is to model the token distribution by ensemble with a disjoint set of chunked contexts $\mathcal{C}=\bigsqcup^K_{k=1} C_k$ to represent $p_\theta(\bm{y}|\mathcal{E},x,t)$ in Eq.~\eqref{eq:mmicl} as 
\begin{gather}     \hat{p}_\theta(\bm{y}|\mathcal{C},x,t)\!\propto\!\prod^K_{k=1} p_\theta(\bm{y}|C_k,x,t)^{w_k},\label{eq:parallel_icl}\\
     C_k=\{e \in \mathcal{E} | \phi(e) = k\},
\end{gather}
where, $\phi: e\to \{1,\dots,K\}$ is a chunking function to partition $\mathcal{E}$ into $\mathcal{C}$.
That is, $\hat{p}_\theta(\bm{y}|\mathcal{C},x,t)$ is defined by an weighted product-of-experts (PoE, \cite{Hinton2002_poe}) of chunk-wise predictive probabilities $p_\theta(\bm{y}|C_k,x,t)$.
In practice, the PoE can be computed by the weighted sum of chunk-wise logits before the softmax for composing next token distributions:
\begin{equation}
    \hat{l}_\theta(y_i)=\sum_{k=1}^{K}w_k l_\theta(y_i|\bm{y}_{<i},C_k,x,t).
\end{equation}
Since each chunked context is shorter than the full context, i.e.,  $|C_k|<|\mathcal{E}|$, we can efficiently compute $l_\theta(\bm{y}|\mathcal{C},x,t)$ by parallel batch-wise forward computations of $p_\theta(\bm{y}|C_k,x,t)$ on devices such as GPUs.
We employ PoE because it is suitable to represent high-dimensional probabilities, such as an LVLM's vocabulary composed of thousands of tokens, as discussed in \cite{Hinton2002_poe}, and a recent study indeed demonstrates that PoE empirically achieves better performance than a mixture-of-experts (MoE, \cite{Jordan_1994_moe}) in the logit-level ensemble reasoning with LLMs~\cite{Chijiwa_arXiv25_lossless_vocab_reduction}.

\begin{figure}[t]
\vspace{-5mm}
\begin{minipage}[t]{\columnwidth}
\begin{algorithm}[H]
    \caption{\proposed}\label{alg:proposed}
    \begin{algorithmic}[1]
        \REQUIRE{Query text $q$, demonstration set $\mathcal{E}$, LLM parameterized by $\theta$, max new token length $L$}
        \ENSURE{Output sequence $\bm{y}$}
        \STATE{$\{C_k\}^K_{k=1}\leftarrow$ Chunking $\mathcal{E}$ by Eq.~\eqref{eq:kmeans_chunking}}
        \STATE{$\{w_k\}^K_{k=1}\leftarrow$ Computing weights by Eq.~\eqref{eq:compilation_weights} on $\mathcal{E}$}
        \STATE{$\bm{y}\leftarrow\emptyset$}
        \FOR{$|\bm{y}|<L$}
        \STATE{$\hat{l}_\theta(y)\!\leftarrow\!\sum_{k=1}^{K}w_k l_\theta(y|\bm{y},C_k,x,t)$\text{\small\color{gray}\# Parallel Decoding}}
        \STATE{$y_\mathrm{text}\sim \operatorname{softmax}(\hat{l}_\theta(y))$}
        \STATE{$\bm{y}\mathrm{.append}(y_\mathrm{text})$ \text{\small\color{gray}\# Add Last Token}}
        \ENDFOR{}
    \end{algorithmic}
\end{algorithm}
\end{minipage}
\vspace{-6mm}
\end{figure}

\subsection{Theoretical Observations}\label{sec:theoretical_observation}
Based on the above formulation, our primary interests are two-fold:
(i) What are desirable properties of chunks by $\phi$ for effective MM-ICL?;
(ii) How should we determine the weights $w_k$ for the ensembles of chunks?
We seek the answers from a theoretical perspective of ensemble learning~\cite{Brown_2009_information_theory_ensemble, Zhou_2010_multi-information_ensemble, Morishita_ICML22_rethinking_fano_ensemble}.
Specifically, based on Fano's inequality~\cite{Fano_1961_transmission_of_information}, the error rate of ensemble predictions can be bounded by:
\begin{theorem}[\citet{Brown_2009_information_theory_ensemble}~and~\citet{Zhou_2010_multi-information_ensemble}]\label{thm_fano}
Given a ground truth $y$, outputs from $K$ models $\mathbf{o}=\{o_1,\dots,o_K\}$, and a reconstruction function $f:\mathbf{o}\mapsto\hat{y}$ (e.g., PoE), the error rate $p_\mathrm{err} = \mathrm{Pr}[y\neq f(\mathbf{o})]$ is bounded by 
\begin{align}
    p_\mathrm{err} > \frac{H(y) - \mathcal{I}(\mathbf{o},y) -1 }{\log_2 |\mathcal{V}|}\label{eq:lower_bound},
\end{align}
where $H$ is entropy and $\mathcal{I}(\mathbf{o}, y)$ is defined as follows:
\begin{align}
    \mathcal{I}(\mathbf{o}, y) &:= I_\mathrm{relev} - I_\mathrm{redun}, \\
    I_\mathrm{relev} &:= \sum^K_{i=1} I(o_i; y),\label{eq:relev} \\
    I_\mathrm{redun} &:= I_\mathrm{multi}(\mathbf{o}|y) - I_\mathrm{multi}(\mathbf{o}), \label{eq:redun}
\end{align}
$I(o;y)$ is mutual information between $o$ and $y$, and $I_\mathrm{multi}(\cdot)$ is called mutual information, which generalizes mutual information for multiple variables.
\end{theorem}
This theorem indicates that the lower bound can be decomposed by the relevance term $I_\mathrm{relev}$ and the redundancy term $I_\mathrm{redun}$ since $H(y)$ and $\log_2 |\mathcal{V}|$ are constants.
Intuitively, $I_\mathrm{relev}$ represents the total correlation between the ground truth $y$ and each model's output $o_i$, i.e., the accuracy of each model's prediction.
Meanwhile, $I_\mathrm{redun}$ can be seen as the indicator of the diversity of the predictions given by $\mathbf{o}$ as the subtraction in Eq.~\eqref{eq:redun} represents the amount of duplicated information among $\mathbf{o}=\{o_1,\dots,o_K\}$.
In summary, this theorem implies that an effective ensemble requires (i) high relevance (i.e., larger $I_\mathrm{relev}$, meaning each model's output $o_i$ is highly correlated with the ground truth $y$, and (ii) high diversity (smaller $I_\mathrm{redun}$), meaning the information shared among models $o$ is minimized.

According to this theoretical observation, (i) selecting chunks based on \textbf{diversity}, and (ii) weighting chunk-wise predictions $\{p_\theta(\bm{y}|C_k,x,t)\}^K_{k=1}$ based on \textbf{task-relevance} are important to improve the performance of \proposed{}.
To achieve these properties, we design (i) context chunking to maximize the diversity of chunk-wise predictions and (ii) context compilation to prioritize the task-relevant predictions.
Since the bounds in Eq.~\eqref{eq:lower_bound} are not computable prior to inference (i.e., before observing $y$), we present practical strategies in the rest of this section.

\subsection{Algorithm}
Algorithm~\ref{alg:proposed} shows the overall procedures of \proposed.

\paragraph*{Context Chunking.}
To ensure the diversity of chunk-wise predictions by $p(y|\mathcal{C}_k, x, t)$, we partition the demonstrations $\mathcal{E}$ into $K$ subsets such that each chunk $C_k$ represents a different type of task.
Concretely, we perform k-means clustering~\cite{Macqueen_1965_kmeans, Lloyd_1982_least_squares_quantization_kmeans} on the demonstration feature $\psi(x,t)$ given by a multi-modal feature extractor $\psi$ (e.g., CLIP~\cite{Radford_ICML21_CLIP, Cherti_CVPR23_openclip}) by solving the following:
\begin{gather}
    \argmin_{\mathcal{C}=\{C_1,\dots,C_K\}} \sum^K_{k=1}\sum^{|C_k|}_{j=1} \|\psi(x^j,t^j)-\bar{\psi}_k\|^2_2, \label{eq:kmeans_chunking}\\
    \psi(x,t) := [\psi_\text{img}(x),\psi_\text{txt}(t)]\label{eq:multimodal_feature}\\
    \bar{\psi}_k = \frac{1}{|C_k|}\sum^{|C_k|}_{j=1} \psi(x^j, t^j),
\end{gather}
where $\psi_\text{img}$ and $\psi_\text{txt}$ are the vision and text encoders of $\psi$.
That is, we expect to increase the relative diversity among chunk-wise predictions by constructing each chunk with similar demonstrations via clustering.
Empirically, we confirm that this context chunking with multi-modal features improves task accuracy and diversity more than other features (i.e., image-/text-only features), as shown in Section~\ref{sec:ex_ablation}.

\paragraph*{Context Compilation.}
To amplify the task-relevance of chunk-wise predictions, we compute a PoE weight $w_k$ in Eq.~\eqref{eq:mmicl} according to the similarity between the task defined by the query question $t$ and the task defined by questions $\{t_1,\dots,t_{|C_k|}\}$ in a demonstration chunk $C_k$:
\begin{align}
    w_k &= \frac{\exp(s_k)}{\sum_{j\in \{1,\dots,K\}} \exp(s_j)}\label{eq:compilation_weights}\\
    s_k &= \frac{1}{|C_k|}\sum^{|C_k|}_{j=1} \frac{\langle \psi(x,t), \psi(x^j,t^j)\rangle}{\|\psi(x,t)\|_2\|\psi(x^j,t^j)\|_2}\
\end{align}
Through this weight computation, we aim to enlarge the overall task-relevance of predictions by prioritizing more task-related predictions in the compilation process.

\begin{figure*}[t]
    \centering
    \begin{minipage}[t]{0.33\linewidth}
      \includegraphics[width=0.85\columnwidth]{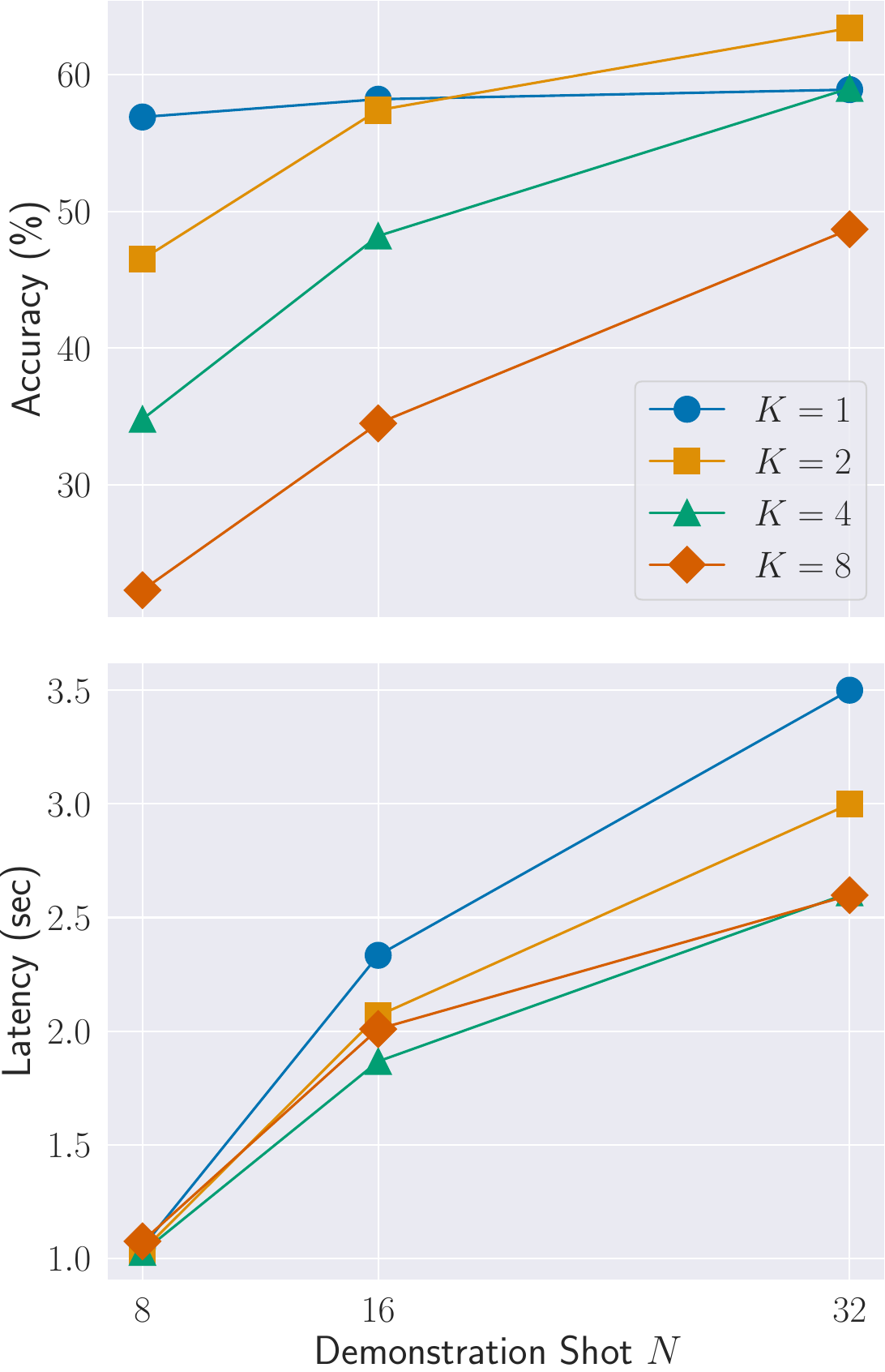}
      \subcaption{LLaVA-OV-7B}
    \end{minipage}
    \begin{minipage}[t]{0.33\linewidth}
      \includegraphics[width=0.85\columnwidth]{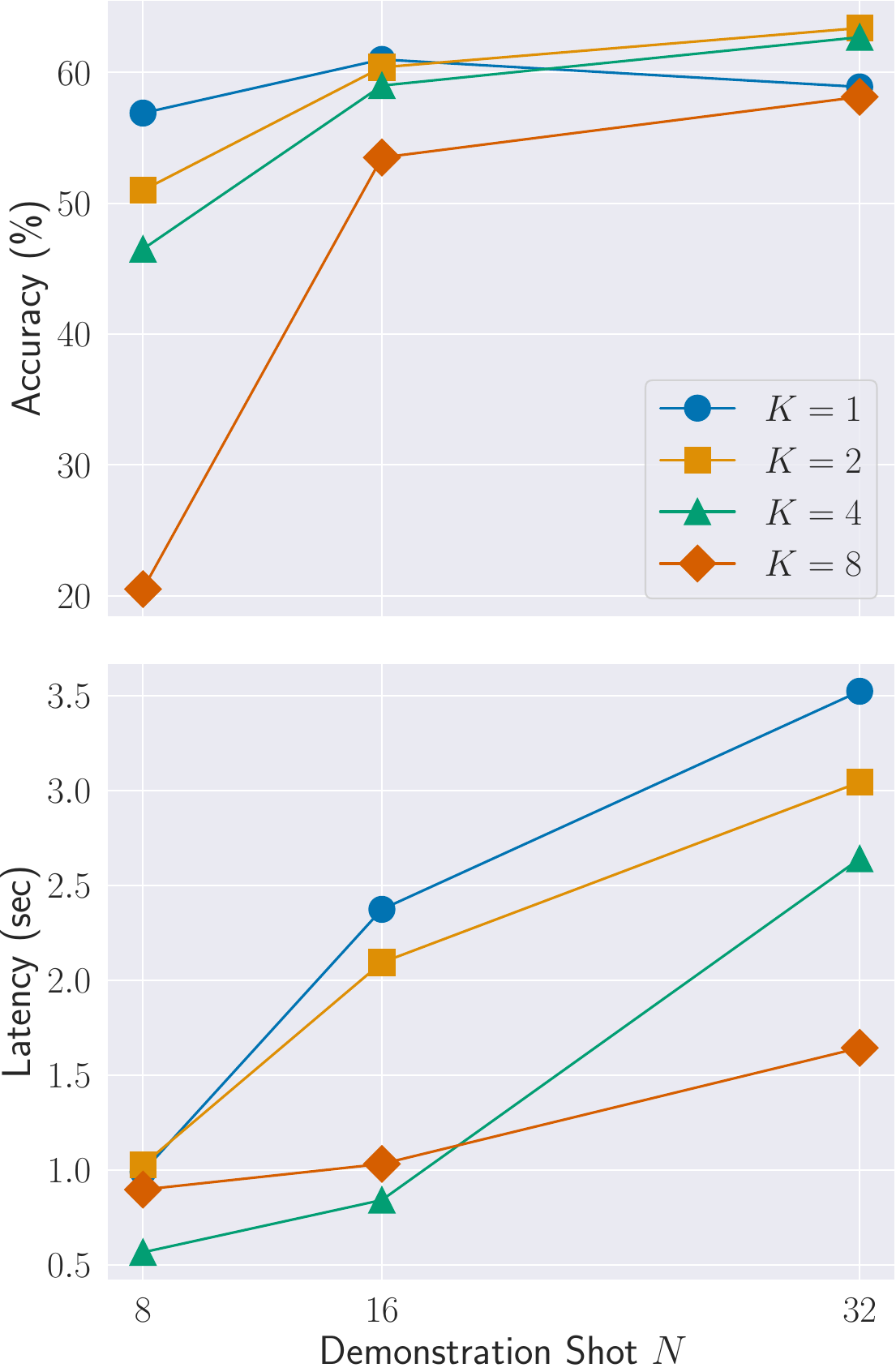}
      \subcaption{Qwen2.5-VL-7B}
    \end{minipage}
    \begin{minipage}[t]{0.33\linewidth}
      \includegraphics[width=0.85\columnwidth]{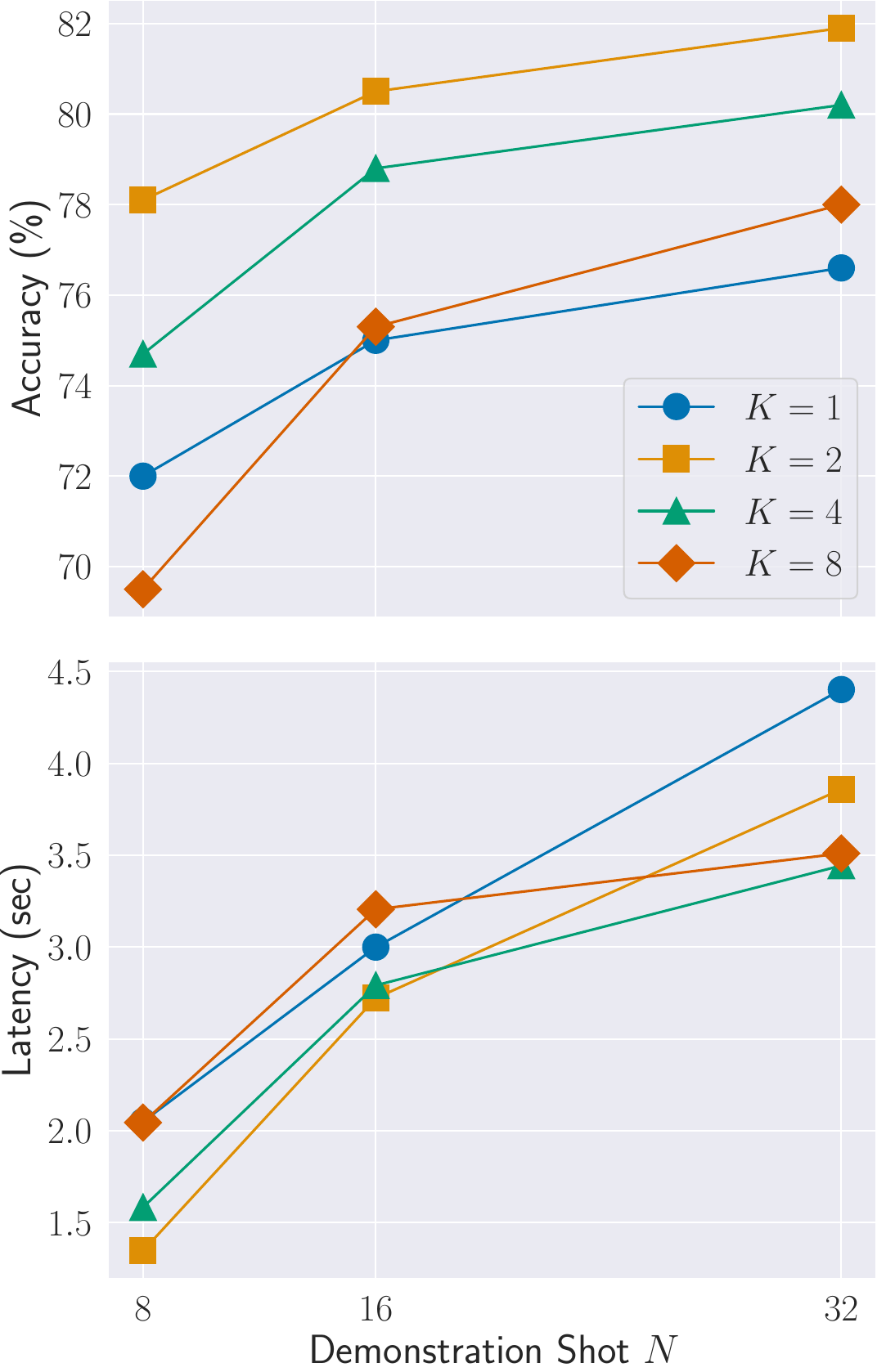}
      \subcaption{InternVL3.5-8B}
    \end{minipage}    
    \caption{
        \textbf{Performance trend of \proposed}. 
        \proposed{} ($K$ > 1) consistently reduces latency across models. Notably, at $N=32$, $K=2$ and $K=4$ can outperform full-context MM-ICL ($K=1$) in accuracy, potentially alleviating the ``lost in the middle'' problem~\cite{Liu_TACL24_lost_in_the_middle}.
    }
    \label{fig:emergence}
    \vspace{-5mm}
\end{figure*}

\section{Experiments}\label{sec:experiments}
We evaluate our \proposed{} through 
(i) a proof-of-concept by varying the demonstration number $N$ and chunk number $K$,
(ii) the inference performance of accuracy and speed across a wide range of benchmarks,
and (iii) the analysis, including ablation studies of \proposed{}.

\subsection{Settings}\label{sec:ex_setting}
\noindent\textbf{Models.}
We used publicly available LVLMs on HuggingFace~\cite{Wolf_arXiv19_huggingface}: LLaVA-OV-7B~\cite{Li_arXiv24_llava-ov}, Qwen-2.5-VL-7B~\cite{Bai_2025_Qwen-vl}, and InternVL3.5 (8B, 14B, 38B)~\cite{Wang_arXiv25_internvl3.5}.

\noindent\textbf{Baselines.}
We compare \proposed{} with two primary plug-and-play methods that do not require additional optimization and architecture-dependent implementations, including
\uline{MM-ICL}: the standard full context inference (equivalent to our method with $K=1$), 
and \uline{DivPrune}~\cite{Alvar_CVPR25_divprune}: a state-of-the-art visual token pruning method.
As DivPrune was not originally designed for MM-ICL, we adapt its core principle to select a \textit{subset} of demonstrations based on diversity, providing a strong baseline for efficient ICL.\looseness-1

\noindent\textbf{Benchmark Datasets.}
We used four diverse multi-modal benchmark datasets in various tasks and domains as follows.
\uline{MI-Bench-ICL} is a specialized benchmark for MM-ICL, composed of closed-ended VQA (C-VQA), open-ended VQA (O-VQA), VQA with error-prone fine-grained visual features (Hallucination), and demo-based task learning (Demo). We primarily used the Demo partition for our proof-of-concept and detailed analysis because it is suitable for examining the ICL capability yielded from demonstrations.
\uline{GQA}~\cite{Hudson_CVPR19_gqa} is a general VQA dataset that contains various questions regarding object/attribute recognition, spatial reasoning, and logical reasoning; we used \texttt{balanced} partitions for testing.
\uline{TextVQA}~\cite{Singh_CVPR19_textvqa} is a VQA specialized for reading texts in images with OCR tokens given by external models. Through TextVQA, we evaluate the capability to leverage OCR tokens to accurately answer questions via MM-ICL.
\uline{COCO Caption}~\cite{Lin_2014_COCO} is a representative image captioning dataset. We tested COCO Caption without specific instructions, i.e., $t^j=\emptyset$, and aimed to make models learn how to generate captions from given demonstration images. 
Except for MI-Bench-ICL, we construct demonstrations for MM-ICL by randomly sampling tuples of (image, question, answer) from the training set.
For MI-Bench-ICL, GQA, and TextVQA, we report accuracy when exact matching outputs with the ground-truth labels.
For COCO Caption, we report the CIDEr score~\cite{Vedantam_CVPR15_cider}.

\noindent\textbf{Evaluation Metrics.}
In addition to task performance scores for each dataset, we report results in several notable evaluation metrics.
To evaluate inference speed, we use \uline{Latency} and \uline{Speedup} metrics, where Latency measures the average inference time in seconds to respond to a single query and Speedup represents the relative improvement ratio of Latency compared to the MM-ICL baseline with the same number of shots. 
\uline{Approximation Ratio} is the relative benchmark performance ratio compared to the original MM-ICL, reported as the average of the ratio for each benchmark performance.
\uline{Diversity} and \uline{Relevance} are the metrics for evaluating whether our method aligns with its theoretical motivation (Section~\ref{sec:theoretical_observation}).
We define proxy metrics for inter-chunk diversity and query relevance based on the KL-divergence as 
\begin{equation}
    \text{Diversity} := \frac{1}{LK}\sum^L_{l=1}\sum^K_{i=1}\sum^K_{j=1, j \neq i} D_\text{KL}(p_i\|p_j),
\end{equation}
\begin{equation}
    \text{Relevance} := \frac{1}{L}\sum^L_{l=1} \exp(-\beta D_\text{KL}(p_\mathcal{E}\|p_\mathcal{C})),
\end{equation}
where $D_\text{KL}(\cdot)$ is KL-divergence, $p_i = p_\theta(y_l|\bm{y}_{<l},C_i,x,t)$, $p_\mathcal{E} = p_\theta(y_l|\bm{y}_{<l},\mathcal{E},x,t)$, $p_\mathcal{C} = \hat{p}_\theta(y_l|\bm{y}_{<l},\mathcal{C},x,t)$, and $\beta$ is a constant; we used fixed $\beta=100$.
Through these metrics, we examine the extent to which \proposed{} achieves the diversity and task-relevance as ensemble distributions demonstrated in Theorem~\ref{thm_fano}.
Note that we used the proxy metrics instead of exact computations of mutual information for Eqs.~\eqref{eq:relev}~and~\eqref{eq:redun} due to the heavy computation cost over a large vocabulary space of LVLMs.

\noindent\textbf{Inference Protocol.}
We used greedy decoding to determine the next token for each iteration, with a maximum new token length of 1024, ensuring that all outputs are produced deterministically.
We varied demonstration number $N$ in $\{8,16,32\}$.
For the chunk number $K$, we used $\{2,4,8\}$, which were determined by using validation sets.
We used up to four NVIDIA H100s with 80GB of memory; .\looseness-1

\begin{table*}[t]
\centering
\caption{\textbf{Performance comparison on multi-modal benchmarks (32-shot)}. Best and second-best scores for each LVLM are \textbf{bolded} and \uline{underlined}.
\proposed{} not only achieves significant speedup, but also yields competitive or superior accuracy (Approx. Ratio > 100\%) compared to full-context MM-ICL. 
Combining \proposed{} with DivPrune yields an excellent accuracy-efficiency trade-off.
}
\label{tb:general_benchmark}
\resizebox{\linewidth}{!}{
\begin{tabular}{@{}l|ccc|ccccccc}
\toprule
 & \multirow{2}{*}{\textbf{Latency}}&\multirow{2}{*}{\textbf{Speedup}} & \multirow{2}{*}{\makecell{\textbf{Approx.} \\ \textbf{Ratio}}} & \multicolumn{4}{c}{\textbf{MI-Bench-ICL}} & \multirow{2}{*}{\textbf{GQA}} & \multirow{2}{*}{\textbf{TextVQA}} & \multirow{2}{*}{\makecell{\textbf{COCO} \\ \textbf{Caption}}} \\
\cmidrule(r){5-8}
& & & & \textbf{C-VQA} & \textbf{O-VQA} & \textbf{Hall.} & \textbf{Demo} & & & \\
\midrule
\multicolumn{11}{@{}l}{\cellcolor{LightBlue}\textbf{LLaVA-OV-7B}} \\
MM-ICL       (32-shot) & 9.662 & 1.000$\times$ & 100.00\% &  \textbf{65.20} & 61.60 & 79.20 & 58.90 & 68.79 & 73.86 & 99.63\\
DivPrune (32$\rightarrow$16-shot) & 5.065 & 1.908$\times$ & 99.63\% & \uline{63.80} & 61.00 & 77.50 & 61.70 & 67.99 & 73.60 & 99.74\\
DivPrune (32$\rightarrow$8-shot) & \textbf{2.834} & \textbf{3.410}$\times$ & 94.39\% & 50.90 & \uline{61.70} & \uline{79.70} & 60.60 & 64.89 & 70.86 & 90.08\\
Parallel-ICL (32-shot, 2-chunk) & 6.673 & 1.448$\times$ & \textbf{100.72}\% & 63.30 & 61.30 & 78.60 & \uline{63.40} & \textbf{68.87} & 74.36 & \uline{101.01} \\
Parallel-ICL (32-shot, 4-chunk) & 5.428 & 1.780$\times$ & 96.51\% & 43.60 & \textbf{61.80} & \textbf{80.70} & 61.00 & 67.80 & \textbf{76.38} & 100.20\\
Parallel-ICL + DivPrune (32$\rightarrow$16-shot, 2-chunk) & \uline{3.486} & \uline{2.772$\times$} & \uline{100.66}\% & 62.90 & \uline{61.70} & 78.90 & \textbf{63.50} & \uline{68.21} & \uline{74.40} & \textbf{100.89}\\
\midrule
\multicolumn{11}{@{}l}{\cellcolor{LightYellow}\textbf{Qwen2.5-VL-7B}} \\
MM-ICL (32-shot) & 2.441 & 1.000$\times$ & \uline{100.00}\% & 65.20 & 61.60 & 79.20 & 58.90 & \textbf{88.02} & \uline{84.28} & \textbf{74.83} \\
DivPrune (32$\rightarrow$16-shot) & 1.219 & 2.002$\times$ & 92.52\% & \uline{66.60} & 60.00 & 75.40 & 51.60 & 69.73 & 82.64 & 67.76 \\
DivPrune (32$\rightarrow$8-shot) & \uline{0.847} & \uline{2.883$\times$} & 78.58\% & 59.00 & 55.10 & 73.40 & 42.80 & 48.78 & 75.22 & 48.04 \\
Parallel-ICL (32-shot, 2-chunk) & 2.115 & 1.154$\times$ & 98.95\% & 63.30 & 61.30 & 78.60 & 63.40 & \uline{84.29} & \textbf{84.38} & \uline{71.39}\\
Parallel-ICL (32-shot, 4-chunk) & 1.642 & 1.486$\times$ & 89.84\% & 52.80 & \uline{61.80} & \uline{80.70} & 62.70 & 65.83 & 79.06 & 60.80 \\
Parallel-ICL + DivPrune (32$\rightarrow$16-shot, 2-chunk) &\textbf{0.814} & \textbf{2.998$\times$} & \textbf{101.90}\% &  \textbf{68.70} & \textbf{62.80} & \textbf{81.60} & \textbf{72.60} & 82.63 & \textbf{84.38} & 69.04 \\
\midrule
\multicolumn{11}{@{}l}{\cellcolor{LightRed}\textbf{InternVL3.5-8B}} \\
MM-ICL (32-shot) & 7.558& 1.000$\times$ & 100.00\% & \uline{84.00}& 61.30& 85.50& 76.60& \textbf{94.94}& 70.12& 89.28\\
DivPrune (32$\rightarrow$16-shot) & 3.840& 1.968$\times$ & 102.02\% & \textbf{84.80}& \textbf{62.50}& 84.80& \uline{80.70}& 91.44& 78.18& 90.66\\
DivPrune (32$\rightarrow$8-shot) & \textbf{2.083}& \textbf{3.628$\times$} & 94.77\% & 74.40& 59.70& \uline{86.30}& 70.60& 90.87& 60.00& 90.47\\
Parallel-ICL (32-shot, 2-chunk) & 5.790& 1.305$\times$ & \textbf{102.61}\% & 81.50& \uline{61.60}& 85.80& \textbf{81.90}& \uline{94.83}& \uline{78.68}& \textbf{92.10}\\
Parallel-ICL (32-shot, 4-chunk) & 5.061& 1.494$\times$ & 101.87\% & 77.70& 62.20& \textbf{86.70}& 80.20& 92.73& \textbf{80.70}& \uline{92.03} \\
Parallel-ICL + DivPrune (32$\rightarrow$16-shot, 2-chunk) & \uline{2.967} & \uline{2.548$\times$} & \uline{102.18}\% & 79.50 & 62.20 & 85.70 & \textbf{81.90} & 94.50 & 78.62 & 91.58 \\
\bottomrule
\end{tabular}
}
\vspace{-5mm}
\end{table*}

\begin{figure}[t]
    \centering
    \includegraphics[width=0.9\linewidth]{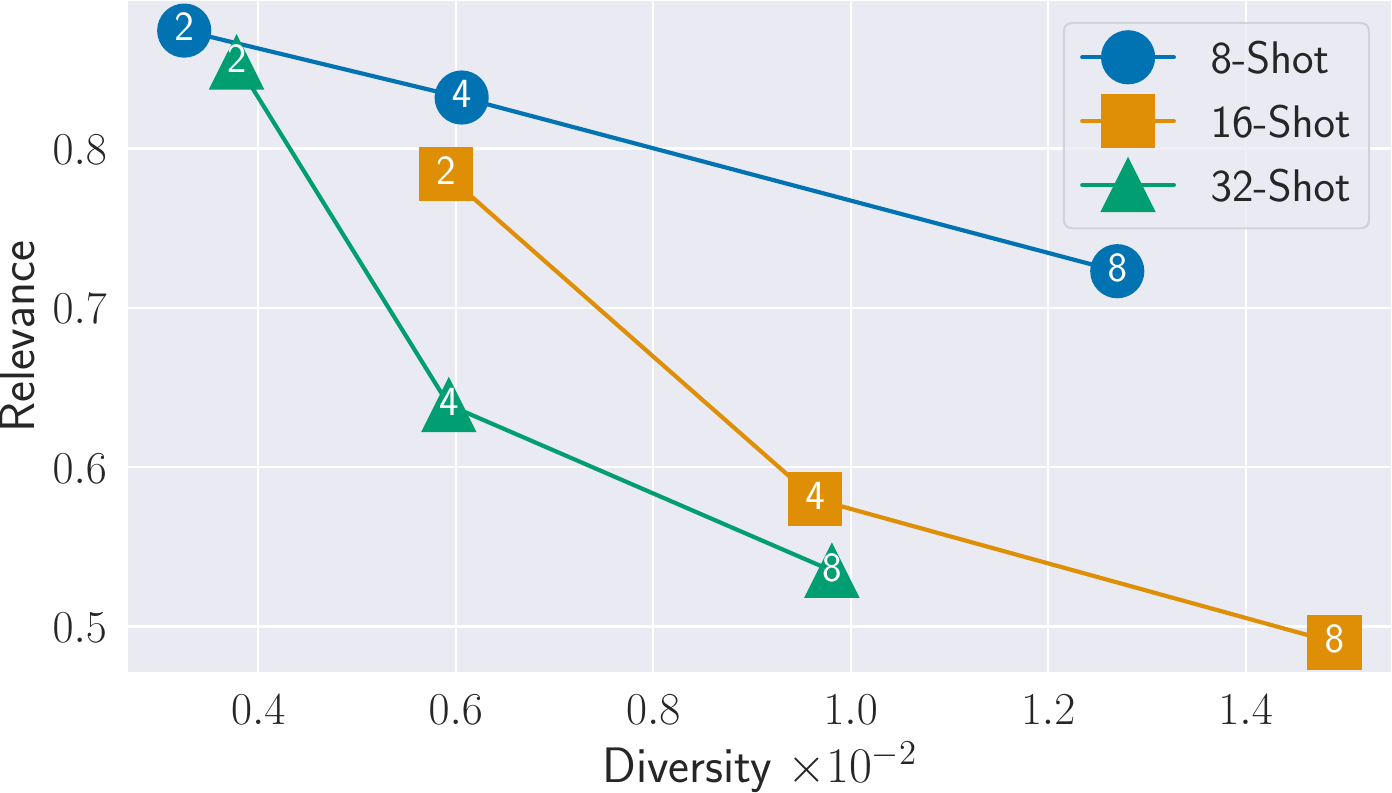}
    \caption{
        \textbf{Diversity-Relevance Tradeoff (chunk number $K$ in marker)}. 
        Increasing the number of chunks improves inter-chunk diversity but decreases the task relevance. This trade-off can be balanced by selecting an appropriate $K$.
    }
    \label{fig:diversity_relevance}
    \vspace{-5mm}
\end{figure}

\subsection{Proof-of-Concept}
We demonstrate a proof-of-concept of \proposed, which approximates MM-ICL by the compilation of predictive distributions from chunked demonstration contexts.
Since MM-ICL processes all demonstrations in a single context, it is non-trivial whether an ensemble of shorter contexts can successfully approximate or even match the full-context performance.
To examine this, we tested \proposed{} on multiple LVLMs (LLaVA-OV-7B, Qwen2.5-VL-7B, and InternVL3.5-8B), by varying the demonstration number $N$ and the chunk number $K$.
Figure~\ref{fig:emergence} shows the trends on MI-Bench-ICL (Demo)~\cite{Liu_EMNLP24_mibench} in the task accuracy and latency, where $K=1$ represents the original MM-ICL.

\noindent\textbf{Task Accuracy.}
The top row of Figure~\ref{fig:emergence} shows that \proposed{} ($K>1$) consistently demonstrates strong ICL capability, with accuracy scaling with $N$,
The accuracy difference between MM-ICL (1-chunk) and \proposed{} diminishes as the demonstration shot increases and the chunk number decreases.
Intriguingly, for all LVLMs at $N=32$, \proposed{} with $K=2$ or $K=4$ consistently outperforms the full-context MM-ICL.
This suggests a remarkable secondary benefit: partitioning the context may mitigate the ``lost in the middle'' phenomenon~\cite{Liu_TACL24_lost_in_the_middle, Baldassini_CVPRW24_what_makes_mmicl}, where models struggle to utilize information buried deep within a long context.
By processing shorter and focused chunks, \proposed{} appears better able to leverage all demonstrations.

\noindent For the chunk number $K$, a larger $K$ tends to yield a lesser approximation of task accuracy for each $N$.
This can be due to the tradeoff between the diversity and relevance of the predictive distributions composed by Eq.~\eqref{eq:parallel_icl}.
As discussed in Section~\ref{sec:theoretical_observation}, both diversity and relevance can contribute to the generalization of ensemble predictions.
In \proposed{}, increasing the number of chunks improves the diversity, while it degrades the relevance as the number of demonstrations included in each chunk context decreases, as shown in Figure~\ref{fig:diversity_relevance}.
Therefore, we recommend selecting an appropriate $K$ that strikes a balance between diversity and relevance.

\noindent\textbf{Inference Latency.}
Secondly, the bottom rows of Figure~\ref{fig:emergence} show that \proposed{} succeeds in reducing the inference latency in almost all cases.
The speedup was particularly significant in large demonstration shots, e.g., $1.32\times$ at $N=32$ on LLaVA-OV-7B.
However, we found that using a larger $K$ does not necessarily improve latency, particularly for smaller $N$, e.g., $(N,K)=(8,8)$.
This is because each chunked context contains an input image $x$ and query text $t$ as defined in Eq.~\eqref{eq:parallel_icl}, and these overheads may outweigh the reduction in latency achieved by parallel inference.
This highlights a trade-off: while large $K$ values are effective for large $N$, the overhead can outweigh the benefit for small $N$.
This limitation can be managed by selecting $K$ based on $N$.

In summary, we observed that \proposed{} can achieve competitive task accuracy while boosting inference speed under some conditions regarding $N$ and $K$, demonstrating a positive answer to our primary research question.

\subsection{Performance on Multiple Benchmarks}
We examine the practicality of \proposed{} through various types of multi-modal benchmarks.
Table~\ref{tb:general_benchmark} summarizes the averaged inference speed and task performance for each benchmark on multiple LVLMs.
As baselines, we also list the results on MM-ICL and DivPrune~\cite{Alvar_CVPR25_divprune}.
Our \proposed{} achieves up to 1.78$\times$ faster inference while maintaining or exceeding the full-context accuracy (Approx. Ratio 96-102\% across models).
Although the demonstration reduction by DivPrune largely speeds up inference, it tends to significantly degrade task performance compared to MM-ICL.
In contrast, \proposed{} can achieve higher performance when the number of demonstrations per chunk matches that of DivPrune, indicating that the performance of ICL with reduced sample size has limitations, and \proposed{} potentially overcomes these limitations while benefiting from speed improvements.
More importantly, Table~\ref{tb:general_benchmark} shows that \proposed{} is orthogonal to demonstration reduction methods. The \proposed{} + DivPrune setting achieves the best overall accuracy-efficiency trade-off (e.g., 3.0$\times$ speedup with 102\% accuracy on Qwen2.5), demonstrating its flexibility.
These results suggest that \proposed{} is an independent and fundamental paradigm with practicality, enabling co-evolution with future improvements in MM-ICL to achieve further performance enhancements and speedups.

\begin{table}[t]
    \centering
    \caption{\textbf{Ablation study on MI-Bench-ICL (Demo) with LLaVA-OV-7B}. Our principled components, k-means-based chunking (vs. random) and similarity-based compilation (vs. uniform), are both crucial for accuracy by improving diversity and relevance.
    }\label{tb:ablation}
    \resizebox{\linewidth}{!}{
    \begin{tabular}{lcccc}\toprule
         \textbf{Method} & \textbf{Accuracy} & \textbf{Latency} & \textbf{Diversity} ($\times10^{-2}$) $\uparrow$ & \textbf{Relevance} $\uparrow$\\\midrule
         MM-ICL  & 58.90 & 3.479 & N/A & 1.000 \\
         Parallel-ICL & 63.40 & 2.999 & 0.378 & 0.854\\
         ~w/ random chunking  & 59.30 & 2.853 & 0.247 & 0.801 \\
         ~w/ uniform compilation & 60.00 & 2.846 & 0.378 & 0.777 \\
         ~w/ textual task features & 58.90 & 2.853 & 0.273 & 0.786 \\
         ~w/ visual task features & 58.20 & 2.936 & 0.310 & 0.778 \\
         \bottomrule
    \end{tabular}
    }
    \vspace{-5mm}
\end{table}

\subsection{Ablation Study}\label{sec:ex_ablation}
We show ablation studies that replace each component of \proposed{} in context chunking, context compilation, and features used for task representations.
Instead of k-means-based chunking by Eq.~\eqref{eq:kmeans_chunking}, we applied random chunking, which selects demonstrations by sampling from a uniform distribution.
For context compilation alternative, we used uniform compilation, ensembling chunk-wise distributions with uniform averaged weights.
We also tried text-only and visual-only features as the task representations for both context chunking and compilation.

Table~\ref{tb:ablation} shows the results on MI-Bench-ICL (Demo) with LLaVA-OV-7B.
Regarding the chunking and compilation algorithm, replacing them with random chunking and uniform compilation largely dropped the diversity and relevance scores, respectively.
These results ensure the design validity of \proposed{}, i.e., enlarging inter-chunk diversity by k-means clustering with Eq.~\eqref{eq:kmeans_chunking} and the task relevance by similarity-based weights Eq.~\eqref{eq:compilation_weights}.
For the task representations, \proposed{} with the multi-modal features by Eq.~\eqref{eq:multimodal_feature} achieved the best performance compared to ones with the uni-modal textual and visual features.
This is consistent with the findings in previous studies, where multi-modal features are more suitable for the task representations~\cite{Alayrac_NeurIPS22_flamingo, Baldassini_CVPRW24_what_makes_mmicl, Qin_NeurIPS24_mmicl_factors}.
Remarkably, all of the alternatives for \proposed{} slightly reduce latency, indicating that the context chunking and compilation can improve performance with little sacrifice to efficiency.

\begin{table}[t]
    \centering
    \caption{
    \textbf{Model scalability of \proposed{} on MI-Bench-ICL (Demo)}.
    \proposed{} consistently improves speed and maintains or enhances the accuracy of MM-ICL, demonstrating its effectiveness across different model scales.
    }\label{tb:model_scalability}
    \resizebox{\linewidth}{!}{
    \begin{tabular}{llccc}\toprule
         \textbf{Model} & \textbf{Method} & \textbf{Accuracy} & \textbf{Latency} & \textbf{Speedup}\\\midrule
         \multirow{3}{*}{\textbf{InternVL3.5-8B}} & MM-ICL (32-shot)  & 76.60 & 4.402 & 1.000$\times$\\
         & Parallel-ICL (32-shot, 2-chunk) & 81.90 & 3.785 & 1.163$\times$\\
         & Parallel-ICL (32-shot, 4-chunk) & 80.20 & 3.338 & 1.319$\times$\\\midrule
         \multirow{3}{*}{\textbf{InternVL3.5-14B}} & MM-ICL (32-shot)  & 83.00 & 6.425 & 1.000$\times$\\
         & Parallel-ICL (32-shot, 2-chunk) & 84.30 & 5.435 & 1.182$\times$\\
         & Parallel-ICL (32-shot, 4-chunk) &84.30 & 4.722 & 1.361$\times$ \\\midrule
         \multirow{3}{*}{\textbf{InternVL3.5-38B}} & MM-ICL (32-shot)  & 86.40 & 20.29 & 1.000$\times$ \\
         & Parallel-ICL (32-shot, 2-chunk) & 87.00 & 17.56 & 1.155$\times$ \\
         & Parallel-ICL (32-shot, 4-chunk) & 87.40 & 15.30 & 1.326$\times$ \\
         \bottomrule
    \end{tabular}
    }
    \vspace{-5mm}
\end{table}

\subsection{Model Size Scalability}
Here, we demonstrate the scalability of \proposed{} regarding LVLM's model sizes.
To this end, we used the 8B, 14B, and 38B models of InternVL3.5.
Table~\ref{tb:model_scalability} shows the results.
For all model sizes, \proposed{} stably improved the inference speed while maintaining or improving the task accuracy.
This indicates that \proposed{} is a flexible inference framework that works with various model sizes.

\section{Conclusion}\label{sec:conclusion}
We propose \proposed{}, a plug-and-play inference algorithm that improves the accuracy-latency trade-off in MM-ICL for LVLMs.
Our method partitions the demonstration context into chunks (context chunking) and processes them in parallel, integrating predictions at the logit level (context compilation).
This principled approach is guided by ensemble learning principles, utilizing clustering to maximize diversity and query similarity for weighted compilation.
Experiments show that \proposed{} significantly improves inference speed (up to 1.78$\times$) while achieving task performance that is comparable to, or even surpasses, that of full-context MM-ICL.
We identify two primary limitations: (i) a query processing overhead that can negate speedups when the number of demonstrations ($N$) is low and the number of chunks ($K$) is high, and (ii) $K$ itself is a crucial hyperparameter that balances the trade-off between inter-chunk diversity and query relevance, requiring careful selection.
Despite these considerations, \proposed{} offers a practical and effective solution to the MM-ICL efficiency challenge.
It also introduces a new paradigm for integrating multiple, diverse contexts, potentially enabling more complex reasoning beyond fixed context lengths.

\clearpage
{\small
\bibliographystyle{ieeenat_fullname}
\bibliography{ref}

\begin{thebibliography}{68}
\providecommand{\natexlab}[1]{#1}
\providecommand{\url}[1]{\texttt{#1}}
\expandafter\ifx\csname urlstyle\endcsname\relax
  \providecommand{\doi}[1]{doi: #1}\else
  \providecommand{\doi}{doi: \begingroup \urlstyle{rm}\Url}\fi

\bibitem[Achiam et~al.(2023)Achiam, Adler, Agarwal, Ahmad, Akkaya, Aleman, Almeida, Altenschmidt, Altman, Anadkat, et~al.]{Achiam_2023_GPT4}
Josh Achiam, Steven Adler, Sandhini Agarwal, Lama Ahmad, Ilge Akkaya, Florencia~Leoni Aleman, Diogo Almeida, Janko Altenschmidt, Sam Altman, Shyamal Anadkat, et~al.
\newblock Gpt-4 technical report.
\newblock \emph{arXiv preprint arXiv:2303.08774}, 2023.

\bibitem[Alayrac et~al.(2022)Alayrac, Donahue, Luc, Miech, Barr, Hasson, Lenc, Mensch, Millican, Reynolds, et~al.]{Alayrac_NeurIPS22_flamingo}
Jean-Baptiste Alayrac, Jeff Donahue, Pauline Luc, Antoine Miech, Iain Barr, Yana Hasson, Karel Lenc, Arthur Mensch, Katherine Millican, Malcolm Reynolds, et~al.
\newblock Flamingo: a visual language model for few-shot learning.
\newblock In \emph{Advances in neural information processing systems}, 2022.

\bibitem[Alvar et~al.(2025)Alvar, Singh, Akbari, and Zhang]{Alvar_CVPR25_divprune}
Saeed~Ranjbar Alvar, Gursimran Singh, Mohammad Akbari, and Yong Zhang.
\newblock Divprune: Diversity-based visual token pruning for large multimodal models.
\newblock In \emph{Proceedings of the Computer Vision and Pattern Recognition Conference}, 2025.

\bibitem[Bai et~al.(2025)Bai, Chen, Liu, Wang, Ge, Song, Dang, Wang, Wang, Tang, et~al.]{Bai_2025_Qwen-vl}
Shuai Bai, Keqin Chen, Xuejing Liu, Jialin Wang, Wenbin Ge, Sibo Song, Kai Dang, Peng Wang, Shijie Wang, Jun Tang, et~al.
\newblock Qwen2. 5-vl technical report.
\newblock \emph{arXiv preprint arXiv:2502.13923}, 2025.

\bibitem[Baldassini et~al.(2024)Baldassini, Shukor, Cord, Soulier, and Piwowarski]{Baldassini_CVPRW24_what_makes_mmicl}
Folco~Bertini Baldassini, Mustafa Shukor, Matthieu Cord, Laure Soulier, and Benjamin Piwowarski.
\newblock What makes multimodal in-context learning work?
\newblock In \emph{Proceedings of the IEEE/CVF Conference on Computer Vision and Pattern Recognition (CVPR) Workshops}, pages 1539--1550, 2024.

\bibitem[Brown(2009)]{Brown_2009_information_theory_ensemble}
Gavin Brown.
\newblock An information theoretic perspective on multiple classifier systems.
\newblock In \emph{International Workshop on Multiple Classifier Systems}, pages 344--353. Springer, 2009.

\bibitem[Brown et~al.(2020)Brown, Mann, Ryder, Subbiah, Kaplan, Dhariwal, Neelakantan, Shyam, Sastry, Askell, Agarwal, Herbert-Voss, Krueger, Henighan, Child, Ramesh, Ziegler, Wu, Winter, Hesse, Chen, Sigler, Litwin, Gray, Chess, Clark, Berner, McCandlish, Radford, Sutskever, and Amodei]{brown_NIPS20_gpt3}
Tom Brown, Benjamin Mann, Nick Ryder, Melanie Subbiah, Jared~D Kaplan, Prafulla Dhariwal, Arvind Neelakantan, Pranav Shyam, Girish Sastry, Amanda Askell, Sandhini Agarwal, Ariel Herbert-Voss, Gretchen Krueger, Tom Henighan, Rewon Child, Aditya Ramesh, Daniel Ziegler, Jeffrey Wu, Clemens Winter, Chris Hesse, Mark Chen, Eric Sigler, Mateusz Litwin, Scott Gray, Benjamin Chess, Jack Clark, Christopher Berner, Sam McCandlish, Alec Radford, Ilya Sutskever, and Dario Amodei.
\newblock Language models are few-shot learners.
\newblock In \emph{Advances in Neural Information Processing Systems}, 2020.

\bibitem[Chen et~al.(2024{\natexlab{a}})Chen, Ye, He, Wang, Khashabi, and Yuille]{Chen_NeurIPS24_llavolta}
Jieneng Chen, Luoxin Ye, Ju He, Zhao-Yang Wang, Daniel Khashabi, and Alan Yuille.
\newblock Efficient large multi-modal models via visual context compression.
\newblock In \emph{Advances in Neural Information Processing Systems}, 2024{\natexlab{a}}.

\bibitem[Chen et~al.(2024{\natexlab{b}})Chen, Zhao, Liu, Bai, Lin, Zhou, and Chang]{Chen_ECCV24_fastv_lvlm_token_pruning}
Liang Chen, Haozhe Zhao, Tianyu Liu, Shuai Bai, Junyang Lin, Chang Zhou, and Baobao Chang.
\newblock An image is worth 1/2 tokens after layer 2: Plug-and-play inference acceleration for large vision-language models.
\newblock In \emph{European Conference on Computer Vision}, 2024{\natexlab{b}}.

\bibitem[Chen et~al.(2025)Chen, Han, He, Liu, Buckley, Qin, Torr, Tresp, and Gu]{Chen_WACV25_truly_mmicl}
Shuo Chen, Zhen Han, Bailan He, Jianzhe Liu, Mark Buckley, Yao Qin, Philip Torr, Volker Tresp, and Jindong Gu.
\newblock Can multimodal large language models truly perform multimodal in-context learning?
\newblock In \emph{2025 IEEE/CVF Winter Conference on Applications of Computer Vision (WACV)}, pages 6000--6010. IEEE, 2025.

\bibitem[Chen et~al.(2024{\natexlab{c}})Chen, Wu, Wang, Su, Chen, Xing, Zhong, Zhang, Zhu, Lu, et~al.]{Chen_CVPR24_internvl}
Zhe Chen, Jiannan Wu, Wenhai Wang, Weijie Su, Guo Chen, Sen Xing, Muyan Zhong, Qinglong Zhang, Xizhou Zhu, Lewei Lu, et~al.
\newblock Internvl: Scaling up vision foundation models and aligning for generic visual-linguistic tasks.
\newblock In \emph{Proceedings of the IEEE/CVF conference on computer vision and pattern recognition}, 2024{\natexlab{c}}.

\bibitem[Cherti et~al.(2023)Cherti, Beaumont, Wightman, Wortsman, Ilharco, Gordon, Schuhmann, Schmidt, and Jitsev]{Cherti_CVPR23_openclip}
Mehdi Cherti, Romain Beaumont, Ross Wightman, Mitchell Wortsman, Gabriel Ilharco, Cade Gordon, Christoph Schuhmann, Ludwig Schmidt, and Jenia Jitsev.
\newblock Reproducible scaling laws for contrastive language-image learning.
\newblock In \emph{Proceedings of the IEEE/CVF Conference on Computer Vision and Pattern Recognition}, pages 2818--2829, 2023.

\bibitem[Chijiwa et~al.(2025)Chijiwa, Hasegawa, Nishida, Yamaguchi, Ohba, Sakao, and Takeuchi]{Chijiwa_arXiv25_lossless_vocab_reduction}
Daiki Chijiwa, Taku Hasegawa, Kyosuke Nishida, Shin'ya Yamaguchi, Tomoya Ohba, Tamao Sakao, and Susumu Takeuchi.
\newblock Lossless vocabulary reduction for auto-regressive language models.
\newblock \emph{arXiv preprint arXiv:2510.08102}, 2025.

\bibitem[Chowdhery et~al.(2023)Chowdhery, Narang, Devlin, Bosma, Mishra, Roberts, Barham, Chung, Sutton, Gehrmann, et~al.]{Chowdhery_JMLR23_palm}
Aakanksha Chowdhery, Sharan Narang, Jacob Devlin, Maarten Bosma, Gaurav Mishra, Adam Roberts, Paul Barham, Hyung~Won Chung, Charles Sutton, Sebastian Gehrmann, et~al.
\newblock Palm: Scaling language modeling with pathways.
\newblock \emph{Journal of Machine Learning Research}, 24\penalty0 (240):\penalty0 1--113, 2023.

\bibitem[Dai et~al.(2023)Dai, Li, Li, Tiong, Zhao, Wang, Li, Fung, and Hoi]{Dai_NeurIPS23_InstructBLIP}
Wenliang Dai, Junnan Li, Dongxu Li, Anthony Tiong, Junqi Zhao, Weisheng Wang, Boyang Li, Pascale Fung, and Steven Hoi.
\newblock Instruct{BLIP}: Towards general-purpose vision-language models with instruction tuning.
\newblock In \emph{Advances in neural information processing systems}, 2023.

\bibitem[Dao(2024)]{Dao_ICLR24_flashattention2}
Tri Dao.
\newblock Flashattention-2: Faster attention with better parallelism and work partitioning.
\newblock In \emph{International Conference on Learning Representations}, 2024.

\bibitem[Fano and Hawkins(1961)]{Fano_1961_transmission_of_information}
Robert~M Fano and David Hawkins.
\newblock Transmission of information: A statistical theory of communications.
\newblock \emph{American Journal of Physics}, 29\penalty0 (11):\penalty0 793--794, 1961.

\bibitem[Gadetsky et~al.(2025)Gadetsky, Atanov, Jiang, Gao, Mighan, Zamir, and Brbic]{Gadetsky_ICLR25_lvlm_unsupervised_icl}
Artyom Gadetsky, Andrei Atanov, Yulun Jiang, Zhitong Gao, Ghazal~Hosseini Mighan, Amir Zamir, and Maria Brbic.
\newblock Large (vision) language models are unsupervised in-context learners.
\newblock In \emph{International Conference on Learning Representations}, 2025.

\bibitem[Gagrani et~al.(2024)Gagrani, Goel, Jeon, Park, Lee, and Lott]{Gagrani_CVPRW24_speculative_decoding_for_lvlm}
Mukul Gagrani, Raghavv Goel, Wonseok Jeon, Junyoung Park, Mingu Lee, and Christopher Lott.
\newblock On speculative decoding for multimodal large language models.
\newblock In \emph{Proceedings of the IEEE/CVF Conference on Computer Vision and Pattern Recognition (CVPR) Workshops}, 2024.

\bibitem[Gao et~al.(2025)Gao, Li, Cao, and Li]{Gao_CVPR25_interleaved-modal_cot}
Jun Gao, Yongqi Li, Ziqiang Cao, and Wenjie Li.
\newblock Interleaved-modal chain-of-thought.
\newblock In \emph{Proceedings of the Computer Vision and Pattern Recognition Conference}, 2025.

\bibitem[Hendel et~al.(2023)Hendel, Geva, and Globerson]{Hendel_EMNLP23_icl_task_vector}
Roee Hendel, Mor Geva, and Amir Globerson.
\newblock In-context learning creates task vectors.
\newblock In \emph{Empirical Methods in Natural Language Processing}, 2023.

\bibitem[Hinton(2002)]{Hinton2002_poe}
Geoffrey~E Hinton.
\newblock Training products of experts by minimizing contrastive divergence.
\newblock \emph{Neural computation}, 14\penalty0 (8):\penalty0 1771--1800, 2002.

\bibitem[Huang et~al.(2025)Huang, Zhai, Shen, Cao, Zhao, Xu, Ye, Hu, and Lin]{Huang_ICLR25_dynamic_llava}
Wenxuan Huang, Zijie Zhai, Yunhang Shen, Shaosheng Cao, Fei Zhao, Xiangfeng Xu, Zheyu Ye, Yao Hu, and Shaohui Lin.
\newblock Dynamic-llava: Efficient multimodal large language models via dynamic vision-language context sparsification.
\newblock In \emph{International Conference on Learning Representations}, 2025.

\bibitem[Hudson and Manning(2019)]{Hudson_CVPR19_gqa}
Drew~A Hudson and Christopher~D Manning.
\newblock Gqa: A new dataset for real-world visual reasoning and compositional question answering.
\newblock In \emph{Proceedings of the IEEE/CVF conference on computer vision and pattern recognition}, 2019.

\bibitem[Ji et~al.(2025)Ji, Zhang, Xia, Chen, Shou, Chen, and Li]{Ji_EMNLP25_specvlm}
Yicheng Ji, Jun Zhang, Heming Xia, Jinpeng Chen, Lidan Shou, Gang Chen, and Huan Li.
\newblock Specvlm: Enhancing speculative decoding of video llms via verifier-guided token pruning.
\newblock In \emph{The 2025 Conference on Empirical Methods in Natural Language Processing}, 2025.

\bibitem[Jiang et~al.(2025)Jiang, Fu, Hao, Hu, Peng, Geng, and Yang]{Jiang_CVPR25_mimic_mmicl}
Yuchu Jiang, Jiale Fu, Chenduo Hao, Xinting Hu, Yingzhe Peng, Xin Geng, and Xu Yang.
\newblock Mimic in-context learning for multimodal tasks.
\newblock In \emph{Proceedings of the IEEE/CVF Conference on Computer Vision and Pattern Recognition}, 2025.

\bibitem[Jordan and Jacobs(1994)]{Jordan_1994_moe}
Michael~I Jordan and Robert~A Jacobs.
\newblock Hierarchical mixtures of experts and the em algorithm.
\newblock \emph{Neural computation}, 6\penalty0 (2):\penalty0 181--214, 1994.

\bibitem[Lauren{\c{c}}on et~al.(2023)Lauren{\c{c}}on, Saulnier, Tronchon, Bekman, Singh, Lozhkov, Wang, Karamcheti, Rush, Kiela, et~al.]{Laurenccon_NeurIPS23_obelics_idefics}
Hugo Lauren{\c{c}}on, Lucile Saulnier, L{\'e}o Tronchon, Stas Bekman, Amanpreet Singh, Anton Lozhkov, Thomas Wang, Siddharth Karamcheti, Alexander Rush, Douwe Kiela, et~al.
\newblock Obelics: An open web-scale filtered dataset of interleaved image-text documents.
\newblock In \emph{Advances in Neural Information Processing Systems}, 2023.

\bibitem[Li et~al.(2024{\natexlab{a}})Li, Zhang, Guo, Zhang, Li, Zhang, Zhang, Zhang, Li, Liu, et~al.]{Li_arXiv24_llava-ov}
Bo Li, Yuanhan Zhang, Dong Guo, Renrui Zhang, Feng Li, Hao Zhang, Kaichen Zhang, Peiyuan Zhang, Yanwei Li, Ziwei Liu, et~al.
\newblock Llava-onevision: Easy visual task transfer.
\newblock \emph{arXiv preprint arXiv:2408.03326}, 2024{\natexlab{a}}.

\bibitem[Li et~al.(2024{\natexlab{b}})Li, Zhang, Zhang, Zhang, Li, Li, Ma, and Li]{Li_2024_llavanext}
Feng Li, Renrui Zhang, Hao Zhang, Yuanhan Zhang, Bo Li, Wei Li, Zejun Ma, and Chunyuan Li.
\newblock Llava-next-interleave: Tackling multi-image, video, and 3d in large multimodal models.
\newblock \emph{arXiv preprint arXiv:2407.07895}, 2024{\natexlab{b}}.

\bibitem[Liang et~al.(2025)Liang, Wang, Xu, Zhou, and Lu]{Liang_CVPR25_efficientllava}
Yinan Liang, Ziwei Wang, Xiuwei Xu, Jie Zhou, and Jiwen Lu.
\newblock Efficientllava: Generalizable auto-pruning for large vision-language models.
\newblock In \emph{Proceedings of the Computer Vision and Pattern Recognition Conference}, 2025.

\bibitem[Lin et~al.(2014)Lin, Maire, Belongie, Hays, Perona, Ramanan, Doll{\'a}r, and Zitnick]{Lin_2014_COCO}
Tsung-Yi Lin, Michael Maire, Serge Belongie, James Hays, Pietro Perona, Deva Ramanan, Piotr Doll{\'a}r, and C~Lawrence Zitnick.
\newblock Microsoft coco: Common objects in context.
\newblock In \emph{European Conference on Computer Vision}. Springer, 2014.

\bibitem[Liu et~al.(2023)Liu, Li, Wu, and Lee]{liu_NeurIPS23_llava}
Haotian Liu, Chunyuan Li, Qingyang Wu, and Yong~Jae Lee.
\newblock Visual instruction tuning.
\newblock In \emph{Advances in Neural Information Processing Systems}, 2023.

\bibitem[Liu et~al.(2024{\natexlab{a}})Liu, Li, Li, and Lee]{Liu_CVPR24_improved_visual_instruction_tuning}
Haotian Liu, Chunyuan Li, Yuheng Li, and Yong~Jae Lee.
\newblock Improved baselines with visual instruction tuning.
\newblock In \emph{Proceedings of the IEEE/CVF Conference on Computer Vision and Pattern Recognition}, 2024{\natexlab{a}}.

\bibitem[Liu et~al.(2024{\natexlab{b}})Liu, Zhang, Xu, Shi, Jiang, Yan, Zhang, Huang, Yuan, Li, et~al.]{Liu_EMNLP24_mibench}
Haowei Liu, Xi Zhang, Haiyang Xu, Yaya Shi, Chaoya Jiang, Ming Yan, Ji Zhang, Fei Huang, Chunfeng Yuan, Bing Li, et~al.
\newblock Mibench: Evaluating multimodal large language models over multiple images.
\newblock In \emph{Proceedings of the 2024 Conference on Empirical Methods in Natural Language Processing}, 2024{\natexlab{b}}.

\bibitem[Liu et~al.(2024{\natexlab{c}})Liu, Lin, Hewitt, Paranjape, Bevilacqua, Petroni, and Liang]{Liu_TACL24_lost_in_the_middle}
Nelson~F Liu, Kevin Lin, John Hewitt, Ashwin Paranjape, Michele Bevilacqua, Fabio Petroni, and Percy Liang.
\newblock Lost in the middle: How language models use long contexts.
\newblock \emph{Transactions of the Association for Computational Linguistics}, 12:\penalty0 157--173, 2024{\natexlab{c}}.

\bibitem[Liu et~al.(2024{\natexlab{d}})Liu, Liu, Wang, Dong, Chen, Rao, Krishna, and Lu]{Liu_ECCV24_elastic_cache}
Zuyan Liu, Benlin Liu, Jiahui Wang, Yuhao Dong, Guangyi Chen, Yongming Rao, Ranjay Krishna, and Jiwen Lu.
\newblock Efficient inference of vision instruction-following models with elastic cache.
\newblock In \emph{European Conference on Computer Vision}, pages 54--69, 2024{\natexlab{d}}.

\bibitem[Lloyd(1982)]{Lloyd_1982_least_squares_quantization_kmeans}
Stuart Lloyd.
\newblock Least squares quantization in pcm.
\newblock \emph{IEEE transactions on information theory}, 28\penalty0 (2):\penalty0 129--137, 1982.

\bibitem[Lu et~al.(2024)Lu, Liu, Zhang, Wang, Dong, Liu, Sun, Ren, Li, Yang, et~al.]{Lu_arXiv24_deepseek-vl}
Haoyu Lu, Wen Liu, Bo Zhang, Bingxuan Wang, Kai Dong, Bo Liu, Jingxiang Sun, Tongzheng Ren, Zhuoshu Li, Hao Yang, et~al.
\newblock Deepseek-vl: towards real-world vision-language understanding.
\newblock \emph{arXiv preprint arXiv:2403.05525}, 2024.

\bibitem[MacQueen(1965)]{Macqueen_1965_kmeans}
James MacQueen.
\newblock Some methods for classification and analysis of multivariate observations [c].
\newblock In \emph{Proc of Berkeley Symposium on Mathematical Statistics \& Probability}, pages 281--297, 1965.

\bibitem[Mitra et~al.(2024)Mitra, Huang, Darrell, and Herzig]{Mitra_CVPR24_CCoT}
Chancharik Mitra, Brandon Huang, Trevor Darrell, and Roei Herzig.
\newblock Compositional chain-of-thought prompting for large multimodal models.
\newblock In \emph{Proceedings of the IEEE/CVF Conference on Computer Vision and Pattern Recognition}, 2024.

\bibitem[Mondal et~al.(2024)Mondal, Modi, Panda, Singh, and Rao]{Mondal_AAAI24_kam-cot}
Debjyoti Mondal, Suraj Modi, Subhadarshi Panda, Rituraj Singh, and Godawari~Sudhakar Rao.
\newblock Kam-cot: Knowledge augmented multimodal chain-of-thoughts reasoning.
\newblock In \emph{Proceedings of the AAAI conference on artificial intelligence}, 2024.

\bibitem[Morishita et~al.(2022)Morishita, Morio, Horiguchi, Ozaki, and Nukaga]{Morishita_ICML22_rethinking_fano_ensemble}
Terufumi Morishita, Gaku Morio, Shota Horiguchi, Hiroaki Ozaki, and Nobuo Nukaga.
\newblock Rethinking fano’s inequality in ensemble learning.
\newblock In \emph{International Conference on Machine Learning}. PMLR, 2022.

\bibitem[Peng et~al.(2024)Peng, chenduo hao, Hu, Peng, Geng, and Yang]{Peng_NeurIPS24_live}
Yingzhe Peng, chenduo hao, Xinting Hu, Jiawei Peng, Xin Geng, and Xu Yang.
\newblock {LIVE}: Learnable in-context vector for visual question answering.
\newblock In \emph{Advances in neural information processing systems}, 2024.

\bibitem[Qin et~al.(2024)Qin, Chen, Fei, Chen, Li, and Che]{Qin_NeurIPS24_mmicl_factors}
Libo Qin, Qiguang Chen, Hao Fei, Zhi Chen, Min Li, and Wanxiang Che.
\newblock What factors affect multi-modal in-context learning? an in-depth exploration.
\newblock In \emph{Advances in Neural Information Processing Systems}, 2024.

\bibitem[Radford et~al.(2021)Radford, Kim, Hallacy, Ramesh, Goh, Agarwal, Sastry, Askell, Mishkin, Clark, et~al.]{Radford_ICML21_CLIP}
Alec Radford, Jong~Wook Kim, Chris Hallacy, Aditya Ramesh, Gabriel Goh, Sandhini Agarwal, Girish Sastry, Amanda Askell, Pamela Mishkin, Jack Clark, et~al.
\newblock Learning transferable visual models from natural language supervision.
\newblock In \emph{International conference on machine learning}. PMLR, 2021.

\bibitem[Shukor et~al.(2024)Shukor, Rame, Dancette, and Cord]{Shukor_ICLR24_beyond_task_perf_mmicl}
Mustafa Shukor, Alexandre Rame, Corentin Dancette, and Matthieu Cord.
\newblock Beyond task performance: evaluating and reducing the flaws of large multimodal models with in-context-learning.
\newblock In \emph{International Conference on Learning Representations}, 2024.

\bibitem[Singh et~al.(2019)Singh, Natarajan, Shah, Jiang, Chen, Batra, Parikh, and Rohrbach]{Singh_CVPR19_textvqa}
Amanpreet Singh, Vivek Natarajan, Meet Shah, Yu Jiang, Xinlei Chen, Dhruv Batra, Devi Parikh, and Marcus Rohrbach.
\newblock Towards vqa models that can read.
\newblock In \emph{Proceedings of the IEEE/CVF conference on computer vision and pattern recognition}, 2019.

\bibitem[Sun et~al.(2024)Sun, Cui, Zhang, Zhang, Yu, Wang, Rao, Liu, Huang, and Wang]{Sun_CVPR24_emu2}
Quan Sun, Yufeng Cui, Xiaosong Zhang, Fan Zhang, Qiying Yu, Yueze Wang, Yongming Rao, Jingjing Liu, Tiejun Huang, and Xinlong Wang.
\newblock Generative multimodal models are in-context learners.
\newblock In \emph{Proceedings of the IEEE/CVF Conference on Computer Vision and Pattern Recognition}, 2024.

\bibitem[Team et~al.(2025)Team, Kamath, Ferret, Pathak, Vieillard, Merhej, Perrin, Matejovicova, Ram{\'e}, Rivi{\`e}re, et~al.]{Google_2025_gemma3}
Gemma Team, Aishwarya Kamath, Johan Ferret, Shreya Pathak, Nino Vieillard, Ramona Merhej, Sarah Perrin, Tatiana Matejovicova, Alexandre Ram{\'e}, Morgane Rivi{\`e}re, et~al.
\newblock Gemma 3 technical report.
\newblock \emph{arXiv preprint arXiv:2503.19786}, 2025.

\bibitem[Tsimpoukelli et~al.(2021)Tsimpoukelli, Menick, Cabi, Eslami, Vinyals, and Hill]{Tsimpoukelli_NeurIPS21_multimodal_few}
Maria Tsimpoukelli, Jacob~L Menick, Serkan Cabi, SM Eslami, Oriol Vinyals, and Felix Hill.
\newblock Multimodal few-shot learning with frozen language models.
\newblock In \emph{Advances in Neural Information Processing Systems}, 2021.

\bibitem[Vaswani et~al.(2017)Vaswani, Shazeer, Parmar, Uszkoreit, Jones, Gomez, Kaiser, and Polosukhin]{Vaswani_NeurIPS17_attention_transformer}
Ashish Vaswani, Noam Shazeer, Niki Parmar, Jakob Uszkoreit, Llion Jones, Aidan~N Gomez, {\L}ukasz Kaiser, and Illia Polosukhin.
\newblock Attention is all you need.
\newblock In \emph{Advances in neural information processing systems}, 2017.

\bibitem[Vedantam et~al.(2015)Vedantam, Lawrence~Zitnick, and Parikh]{Vedantam_CVPR15_cider}
Ramakrishna Vedantam, C Lawrence~Zitnick, and Devi Parikh.
\newblock Cider: Consensus-based image description evaluation.
\newblock In \emph{Proceedings of the IEEE conference on computer vision and pattern recognition}, pages 4566--4575, 2015.

\bibitem[Wang et~al.(2025)Wang, Gao, Gu, Pu, Cui, Wei, Liu, Jing, Ye, Shao, et~al.]{Wang_arXiv25_internvl3.5}
Weiyun Wang, Zhangwei Gao, Lixin Gu, Hengjun Pu, Long Cui, Xingguang Wei, Zhaoyang Liu, Linglin Jing, Shenglong Ye, Jie Shao, et~al.
\newblock Internvl3. 5: Advancing open-source multimodal models in versatility, reasoning, and efficiency.
\newblock \emph{arXiv preprint arXiv:2508.18265}, 2025.

\bibitem[Wolf et~al.(2019)Wolf, Debut, Sanh, Chaumond, Delangue, Moi, Cistac, Rault, Louf, Funtowicz, et~al.]{Wolf_arXiv19_huggingface}
Thomas Wolf, Lysandre Debut, Victor Sanh, Julien Chaumond, Clement Delangue, Anthony Moi, Pierric Cistac, Tim Rault, R{\'e}mi Louf, Morgan Funtowicz, et~al.
\newblock Huggingface's transformers: State-of-the-art natural language processing.
\newblock \emph{arXiv preprint arXiv:1910.03771}, 2019.

\bibitem[Wu et~al.(2024)Wu, Sun, Li, Welleck, and Yang]{Wu_NeurIPS24_scaling_inference_computation_llm}
Yangzhen Wu, Zhiqing Sun, Shanda Li, Sean Welleck, and Yiming Yang.
\newblock Scaling inference computation: Compute-optimal inference for problem-solving with language models.
\newblock In \emph{The 4th Workshop on Mathematical Reasoning and AI at NeurIPS'24}, 2024.

\bibitem[Xing et~al.(2025)Xing, Huang, Dong, Lu, Zhang, Zang, Cao, He, Wang, Wu, et~al.]{Xing_CVPR25_pyramid_drop}
Long Xing, Qidong Huang, Xiaoyi Dong, Jiajie Lu, Pan Zhang, Yuhang Zang, Yuhang Cao, Conghui He, Jiaqi Wang, Feng Wu, et~al.
\newblock Conical visual concentration for efficient large vision-language models.
\newblock In \emph{Proceedings of the Computer Vision and Pattern Recognition Conference}, 2025.

\bibitem[Xu et~al.(2025)Xu, Jin, Wu, Li, Song, Sun, and Yuan]{Xu_ICCV25_llava-cot}
Guowei Xu, Peng Jin, Ziang Wu, Hao Li, Yibing Song, Lichao Sun, and Li Yuan.
\newblock Llava-cot: Let vision language models reason step-by-step.
\newblock In \emph{Proceedings of the International Conference on Computer Vision}, 2025.

\bibitem[Yang et~al.(2025)Yang, Dong, Zhu, Su, Wang, Tian, Chen, Wang, Lu, and Dai]{Yang_CVPR25_pvc}
Chenyu Yang, Xuan Dong, Xizhou Zhu, Weijie Su, Jiahao Wang, Hao Tian, Zhe Chen, Wenhai Wang, Lewei Lu, and Jifeng Dai.
\newblock Pvc: Progressive visual token compression for unified image and video processing in large vision-language models.
\newblock In \emph{Proceedings of the Computer Vision and Pattern Recognition Conference}, 2025.

\bibitem[Zhang et~al.(2024{\natexlab{a}})Zhang, Sun, Chen, Xiao, Shao, Zhang, Liu, Chen, and Luo]{Zhang_ECCV24_gpt4roi}
Shilong Zhang, Peize Sun, Shoufa Chen, Min Xiao, Wenqi Shao, Wenwei Zhang, Yu Liu, Kai Chen, and Ping Luo.
\newblock Gpt4roi: Instruction tuning large language model on region-of-interest.
\newblock In \emph{European conference on computer vision}, 2024{\natexlab{a}}.

\bibitem[Zhang et~al.(2025{\natexlab{a}})Zhang, Fang, Yang, and Feng]{Zhang_ICLR25_llava-mini}
Shaolei Zhang, Qingkai Fang, Zhe Yang, and Yang Feng.
\newblock Llava-mini: Efficient image and video large multimodal models with one vision token.
\newblock In \emph{International Conference on Learning Representations}, 2025{\natexlab{a}}.

\bibitem[Zhang et~al.(2025{\natexlab{b}})Zhang, Li, Chu, Xu, Yang, Guan, Xu, Jing, Cui, et~al.]{Zhang_CVPR25_lvlm_ood}
Xingxuan Zhang, Jiansheng Li, Wenjing Chu, Renzhe Xu, Yuqing Yang, Shikai Guan, Jiazheng Xu, Liping Jing, Peng Cui, et~al.
\newblock On the out-of-distribution generalization of large multimodal models.
\newblock In \emph{Proceedings of the IEEE/CVF Conference on Computer Vision and Pattern Recognition}, 2025{\natexlab{b}}.

\bibitem[Zhang et~al.(2025{\natexlab{c}})Zhang, Fan, Ma, Zheng, Huang, Cheng, Gudovskiy, Okuno, Nakata, Keutzer, et~al.]{Zhang_ICML25_sparsevlm}
Yuan Zhang, Chun-Kai Fan, Junpeng Ma, Wenzhao Zheng, Tao Huang, Kuan Cheng, Denis~A Gudovskiy, Tomoyuki Okuno, Yohei Nakata, Kurt Keutzer, et~al.
\newblock Sparsevlm: Visual token sparsification for efficient vision-language model inference.
\newblock In \emph{International Conference on Machine Learning}, 2025{\natexlab{c}}.

\bibitem[Zhang et~al.(2024{\natexlab{b}})Zhang, Zhang, Li, hai zhao, Karypis, and Smola]{Zhang_TMLR_mm-cot}
Zhuosheng Zhang, Aston Zhang, Mu Li, hai zhao, George Karypis, and Alex Smola.
\newblock Multimodal chain-of-thought reasoning in language models.
\newblock \emph{Transactions on Machine Learning Research}, 2024{\natexlab{b}}.

\bibitem[Zhao et~al.(2024)Zhao, Cai, Si, Ma, An, Chen, Liu, Wang, Han, and Chang]{Zhao_ICLR24_mmicl}
Haozhe Zhao, Zefan Cai, Shuzheng Si, Xiaojian Ma, Kaikai An, Liang Chen, Zixuan Liu, Sheng Wang, Wenjuan Han, and Baobao Chang.
\newblock Mmicl: Empowering vision-language model with multi-modal in-context learning.
\newblock In \emph{International Conference on Learning Representations}, 2024.

\bibitem[Zhou and Li(2010)]{Zhou_2010_multi-information_ensemble}
Zhi-Hua Zhou and Nan Li.
\newblock Multi-information ensemble diversity.
\newblock In \emph{International Workshop on Multiple Classifier Systems}, pages 134--144. Springer, 2010.

\bibitem[Zhuang et~al.(2025)Zhuang, Singh, Liu, Shang, and Gao]{Zhuang_ICLR25_vectoricl}
Yufan Zhuang, Chandan Singh, Liyuan Liu, Jingbo Shang, and Jianfeng Gao.
\newblock Vector-{ICL}: In-context learning with continuous vector representations.
\newblock In \emph{International Conference on Learning Representations}, 2025.

\bibitem[Zong et~al.(2025)Zong, Bohdal, and Hospedales]{Zong_ICLR2025_vl-icl_bench}
Yongshuo Zong, Ondrej Bohdal, and Timothy Hospedales.
\newblock Vl-icl bench: The devil in the details of multimodal in-context learning.
\newblock In \emph{The Thirteenth International Conference on Learning Representations}, 2025.

\end{thebibliography}
}



\end{document}